\documentclass[twoside,11pt]{article}

\usepackage{jmlr2e}
\usepackage{colortbl}
\usepackage{siunitx}
\usepackage{booktabs}
\usepackage[font=footnotesize,labelfont=it,tableposition=top]{caption}
\usepackage{subcaption}
\usepackage{float}
\usepackage{mathtools}
\usepackage[capitalize,nameinlink,]{cleveref} %
\usepackage[linesnumbered,ruled,vlined]{algorithm2e}
\DontPrintSemicolon
\SetKwInput{KwInput}{Input}
\SetKwInput{KwOutput}{Output}
\SetKwComment{Comment}{\# }{}
\usepackage{setspace}
\crefname{section}{\S}{\S\S}
\Crefname{section}{\S}{\S\S}
\crefname{algocf}{Alg.}{Algs.}
\Crefname{algocf}{Algorithm}{Algorithms}

\definecolor{shadecolor}{gray}{0.9}

\jmlrheading{1}{}{}{}{}{}

\DeclarePairedDelimiterX{\infdivx}[2]{(}{)}{%
#1\;\delimsize\|\;#2%
}

\newcommand{\KLnew}{\ensuremath{\textsc{kl}}\infdivx}

\ShortHeadings{Class Balanced Dynamic Acquisition}{Schachtsiek, Rossi and Hannagan}
\firstpageno{1}

\begin{document}

\title{Class Balanced Dynamic Acquisition for Domain Adaptive Semantic Segmentation using Active Learning}

\author{\name Marc Schachtsiek \email marc.schachtsiek@dauphine.eu \\
        \addr PSL University \& Stellantis \email marc.schachtsiek@gmail.com\footnotemark[1] \\
        \name Simone Rossi \email simone.rossi2@stellantis.com \\
        \addr Stellantis \\
        \name Thomas Hannagan \email thomas.hannagan@stellantis.com \\
        \addr Stellantis \\}

\maketitle

\footnotetext[1]{Please reach out via this email address.}

\begin{abstract}%
Domain adaptive active learning is leading the charge in label-efficient training of neural networks. 
For semantic segmentation, state-of-the-art models jointly use two criteria of uncertainty and diversity to select training labels, combined with a pixel-wise acquisition strategy. 
However, we show that such methods currently suffer from a class imbalance issue which degrades their performance for larger active learning budgets. 
We then introduce Class Balanced Dynamic Acquisition (CBDA), a novel active learning method that mitigates this issue, especially in high-budget regimes. 
The more balanced labels increase minority class performance, which in turn allows the model to outperform the previous baseline by 0.6, 1.7, and 2.4 mIoU for budgets of 5\%, 10\%, and 20\%, respectively. 
Additionally, the focus on minority classes leads to improvements of the minimum class performance of $0.5$, $2.9$, and $4.6$ IoU respectively.
The top-performing model even exceeds the fully supervised baseline, showing that a more balanced label than the entire ground truth can be beneficial.
\end{abstract}


\section{Introduction}

Semantic segmentation is a key computer vision task with a wide variety of applications in autonomous driving \citep{autodr2,autodr1}, medical analysis \citep{medical1,unet}, and other domains \citep{other1}.
Training semantic segmentation models typically relies on a large amount of labeled data.
Since a lot of this data, i.e. real-life autonomous driving data, needs to be labeled by human annotators, it is expensive.
Maximizing model performance with as few human labels as possible is a key research area, both to improve semantic segmentation models and training in general and to move closer to production feasibility.

Domain Adaptation and Active Learning (AL) are two techniques that are used to reduce the number of labels required.
Domain Adaptation uses an out-of-domain source and in-domain target dataset.
The source data typically consists of samples that can easily be acquired and automatically labeled - for example, from a simulation.
The target data are labeled samples from the target domain.
Active Learning seeks to determine the most beneficial samples to be labeled instead of labeling all of them.
Generally, heuristics based on the model predictions are used for that determination.
These two methods can be combined for even more label efficient training.

Another aspect of deep neural networks for autonomous driving more broadly is safety and reliability.
The most critical aspect to evaluate a model for such environments is the lowest performance of any of its parts.
Even if meeting the requirements in a majority of cases, incorrect predictions in an untypical driving scenario can have unpredictable effects.
Protecting vulnerable members of traffic is critical to meeting safety requirements.
Given that less than $1\%$ of the labels for Cityscapes for example cover bike, motorcycle and rider, there is a data imbalance problem.
This problem is shared across any setup, but active learning provides a means to address it.

\begin{figure}[t]
    \centering
    \includegraphics[width=0.95\textwidth]{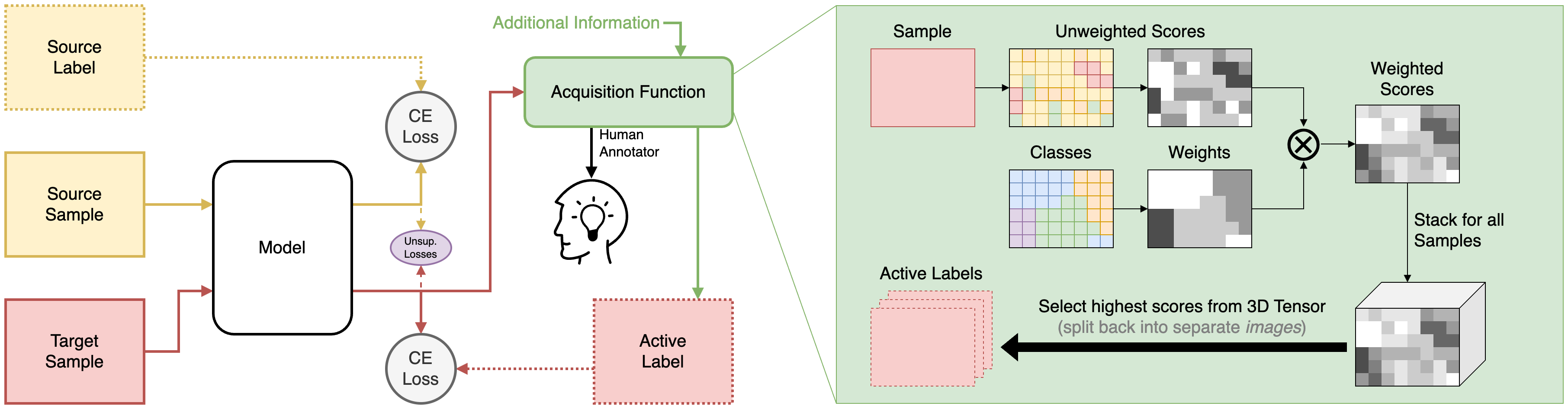}
    \caption{\textbf{Overview of Class Balanced Dynamic Acquisition (CBDA)}. Left: the generic setup of active learning for domain adaptation relies crucially on an acquisition function whereby pixels receive scores that determine whether a label will be queried. Right: We introduce CBDA as an acquisition function that weighs scores according to observed class statistics and shares the labeling budget flexibly across samples.}
    \label{fig:overview}
\end{figure}

\paragraph{Contributions.} 
In the context of domain adaptation with active learning, we analyze the effect of imbalanced data for semantic segmentation in an autonomous driving scenario and we introduce a novel method, CBDA---Class Balanced Dynamic Acquisition---to obtain more balanced active labels for training (see \cref{fig:overview} for an overview).
Notably, this new approach does not increase the computational complexity and can be used as a drop-in replacement in the standard training pipeline. 
We analyze the benefits of this approach to improve sample efficiency on GTAV $\rightarrow$ Cityscapes, showing improvements in overall performance by increasing minority class scores, without being detrimental for other classes.

\subsection{Setup}

The setup of domain adaptive semantic segmentation with active learning consists of two datasets, the labeled source dataset $ \mathcal{S}=\{(X_s,Y_s)\}$ and the unlabeled target dataset $\mathcal{T}=\{(X_t)\}$, and the model and training objectives.
The final model is evaluated on a target test set.
The training setup follows the iterative active learning strategy from \citep{RIPU} where active learning is performed at specific training iterations.

First, the model $\mathcal{M}$ is trained using $X_s$, $Y_s$, and $X_t$ with a supervised cross-entropy loss for the source data and some unsupervised objectives for the target samples.
Once an active learning iteration is reached each image $I$'s predicted class probabilities $P=softmax(\mathcal{M}(I))$ and pseudo label $\hat{Y}=\arg \max_{c \in \{1,\ldots,C\}} P_c$ are calculated.
Both are pixel-wise outputs and pixel at position $i$,$j$ can be accessed via either $P^{(i,j)}$ or $\hat{Y}^{(i,j)}$.

The acquisition function determines which sample pixels are selected for labeling and make up the target active label $\tilde{Y}_t$.
This typically takes the form of a pixel-wise acquisition score matrix $\mathcal{A}$ that is calculated for each image using the class probabilities and pseudo label and the indices with the highest scores are selected according to an active learning budget.
These indices correspond to the coordinates of the pixel in the respective image.
During a real-world application, these samples would be given to human annotators for annotation, but for this experimental setup, the already existing ground truth target labels $Y_t$ are used.
For each selected pixel, the corresponding ground truth class is selectively provided to the model for training.

In the training iterations following the active learning iteration, the target active labels are incorporated in the standard training setup and serve as labels for the target samples $X_t$.
They are trained with the same supervised cross-entropy loss as the source sample-label pairs, ignoring the loss of each pixel that is not part of the active label.
Pixels that have already been selected are ignored.

\paragraph{Acquisition functions.}

In the image-based AL scenario, the acquisition scores are averaged across each image in the target set and the highest-scoring images are selected for labeling.
The averaging of the scores loses any pixel-wise information.
\cref{alg:ilm} gives an overview of image-wise active learning.

The pixel-based approaches keep the score for each pixel, so each image has a resulting acquisition score matrix $\mathcal{A}^{(i,j)} \in \mathbb{R}^{H \times W}$.
However, picking the highest scoring pixels for labeling is then done iteratively for each image once its acquisition score matrix has been calculated as shown in \cref{alg:ripu}.
It is split equally across all the images to stick to the AL budget.
The specific pixel-wise acquisition function that serves as the baseline for the experiments is "Region Acquisition" or RA \citep{RIPU}.

\begin{minipage}[t]{0.46\textwidth}
    \begin{algorithm}[H]
        \footnotesize
        \caption{Image-Wise Acquisition}
        \label{alg:ilm}
        \vspace{5pt}
        $scores =$ list()\;
        \ForEach{$I$ \textup{\textbf{in}} $\{X_t\}$}
        {
            $\mathcal{A} = calculate\_scores(I)$\;
            $scores$.$append(\mathcal{A}$.$mean())$\;
        }
        $selection = select\_images(scores, iter\_budget)$\;
        \ForEach{$I$ \textup{\textbf{in}} $selection$}
        {
            $\tilde{Y}_I = Y_I$\;
        }
    \end{algorithm}
\end{minipage}
\hfill
\begin{minipage}[t]{0.46\textwidth}
    \begin{algorithm}[H]
        \footnotesize
        \caption{Pixel-Wise Acquisition}
        \label{alg:ripu}
        \vspace{5pt}
        \ForEach{$I$ \textup{\textbf{in}} $\{X_t\}$}
        {
            $\mathcal{A} = calculate\_scores(I)$\;
            $selection = select\_pixels(\mathcal{A}, \frac{iter\_budget}{|\{X_t\}|})$\;
            \ForEach{$(i,j)$ \textup{\textbf{in}} $selection$}
            {
                $\tilde{Y}_I^{(i,j)} = Y_I^{(i,j)}$\;
            }
        }
    \end{algorithm}
\end{minipage}

\paragraph{Related work.}

\citet{RIPU} claimed the class balancing effect of their method as a beneficial factor.
However, when examining higher budget regimes exceeding 5\% the active label becomes progressively more imbalanced, at over 10\% even more imbalanced than the entire ground truth data.
This coincides with progressively worse model performance for these high budgets.
Given the progressively worsening imbalance and the relatively lower scores of the minority classes, actively balancing the target labels that were selected was an obvious next step.
There are previous methods in the space of class balanced active learning \citep{cbal-static,cbal-optim}, yet none have been applied to semantic segmentation specifically.
\citet{cbal-static} propose a method that selects the highest scoring samples within some budget per class.
This budget is inflexible and even if samples from another class have significantly lower acquisition scores the budget can not be used for or at least partially reallocated to a class with higher acquisition scores.
The method by \citet{cbal-optim} uses an additional optimization problem to balance the selection; however, considering each pixel a sample for semantic segmentation makes their method prohibitively expensive - either in memory for the required matrices, or in computational complexity if they are processed in a streaming fashion.

\section{Method}

The proposed method introduces a novel mechanism to select a more balanced active label during training that can be applied to pixel-based semantic segmentation with active learning.
This takes the form of class weights applied to the acquisition score matrices based on statistics of previously gathered active labels.
\cref{fig:overview} shows a method overview.

\paragraph{Dynamic Acquisition (DA).}

The two previously presented approaches of image- and pixel-based acquisition can be combined to keep the advantages of both; this new acquisition function is called \textit{Dynamic Acquisition (DA)}.
Instead of calculating the score and picking the pixels immediately after for each image, the entire set of acquisition score matrices is calculated, then stacked together and the pixels with the highest scores are selected from that tensor $\mathcal{A}_S \in \mathbb{R}^{|\{X_t\}| \times H \times W}$.

\cref{sec:da_usage} shows the distribution of the percentage of selected pixels of DA vs RA, showing that the model utilizes the budget more dynamically and selects a wider variety.

\paragraph{Class Balancing (CB).}

Our novel class balancing method calculates a class budget similarly to \citet{cbal-static} but instead of applying the remaining class budget statically, it is used to calculate a class weight that downweighs the acquisition scores of classes in proportion to how many samples of that class have been selected in previous AL iterations compared to a goal distribution. %
In a realistic setting no prior information about the ground truth class distribution would be available, therefore a uniform distribution is the most sensible target.
If under specific circumstances extra information would be available, the goal distribution could be chosen differently.
Notably, this goal is not applied strictly, but is effectively used as a loose target.

Like other methods, ours relies on computing statistics of the classes of the set of active labels $\{\tilde{Y}_t^{(i,j)}\}$.
For each class $c \in \{1,\ldots,C\}$ the cumulative number of labels of that class $L_{c,i}$ are being counted for the current iteration.
To determine the class weight, the first step is to calculate the ideal iteration class budget $B_{c,i}$ for each class for the current AL iteration as shown in \cref{eqn:class_budget}.
This part relates to the goal distribution and can be modified to calculate a budget matching any goal.
The equation uses the total number of pixels of the target samples, the percentage active learning budget $B_{AL}$, the number of active learning iterations $N_i$, the current iteration index $i \in \{1,\ldots,N_i\}$ and the number of classes $C$.
\begin{equation}
    \label{eqn:class_budget}
    B_{c,i} = (|\mathcal{T}| \times W \times H) \times \frac{B_{AL}}{C} \times \frac{i}{N_i}
\end{equation}
The class weight for the given iteration $W_{c,i}$ is then calculated as shown in \cref{eqn:weight}. 
The maximum function is being used to prevent the weight from reaching exactly 0, with $\epsilon$ being set to a small value strictly larger than 0 - for example, $\epsilon = \num{1e-6}$. %
This serves the purpose of keeping the relative differences between the acquisition scores for labels of classes that have filled or exceeded the budget.
\begin{equation}
    \label{eqn:weight}
    W_{c,i} = \max \left(1 - \frac{L_{c,i}}{B_{c,i}}, \epsilon \right)
\end{equation}
Finally, in every AL iteration the weighted acquisition scores $\hat{\mathcal{A}}$ are calculated by multiplying the acquisition scores with the weight for the class as predicted by the pseudo label as shown in \cref{eqn:apply_weight}. 
Since the weight is in the range $[\epsilon,1]$, it downweighs the classes that have filled or exceeded their budget. 
This will reduce the chance of them getting selected by the acquisition function but will not remove the possibility entirely contrary to the method by \citet{cbal-static}.
\begin{equation}
    \label{eqn:apply_weight}
    \hat{\mathcal{A}}^{(i,j)} = \mathcal{A}^{(i,j)} \times W_{c} \; \textrm{where} \, \hat{Y}_t^{(i,j)} = c
\end{equation}
Combining the previous equations and method results in the algorithm presented in \cref{sec:cbda}. 
In this case, it is combined with the dynamic acquisition strategy presented in the previous section for best results, but the same method can be combined with any active learning strategy and is heuristics agnostic.

\section{Results}

\cref{tab:results_gtav} shows the results for the GTAV $\rightarrow$ Cityscapes tasks.
The results using SYNTHIA as the source dataset are shown in \cref{sec:results_synthia}.
These specific dataset combinations were selected for best comparability with previous approaches.
The experiments between the double lines are re-runs of the baseline method and of the proposed method, all others were taken from the respective papers \citep{RIPU,AADA,MADA,ILM-ASSL}.
They are all based on the DeepLabv-3+ architecture \citep{deeplabv3+} with a ResNet-101 backbone \citep{resnet}.
Contrary to previous papers, the classes listed here are not in order of the internal train id but are sorted in descending order of the class frequency according to the ground truth label distribution.

\begin{table}[ht]
    \tiny
    \centering
    \caption{Results of the GTAV $\rightarrow$ Cityscapes task}
    \label{tab:results_gtav}
    \setlength{\tabcolsep}{2pt}
    \begin{tabular}
    {
        @{}
        lcccccccccccccccccccc
        @{}
    }
        \toprule
        Method                          
        & \rotatebox{60}{road} 
        & \rotatebox{60}{buil.}
        & \rotatebox{60}{veg.}
        & \rotatebox{60}{car}
        & \rotatebox{60}{side.}
        & \rotatebox{60}{sky}
        & \rotatebox{60}{pers.}
        & \rotatebox{60}{pole}
        & \rotatebox{60}{terr.}
        & \rotatebox{60}{fence}
        & \rotatebox{60}{wall}
        & \rotatebox{60}{sign}
        & \rotatebox{60}{bike}
        & \rotatebox{60}{train}
        & \rotatebox{60}{truck}
        & \rotatebox{60}{bus}
        & \rotatebox{60}{light}
        & \rotatebox{60}{rider}
        & \rotatebox{60}{motor.}
        & mIoU \\ \midrule
        Source only & 75.8 & 77.2 & 81.3 & 49.9 & 16.8 & 70.3 & 53.8 & 25.5 & 24.6 & 21.0 
                    & 12.5 & 20.1 & 36.0 &  6.5 & 17.2 & 25.9 & 30.1 & 26.4 & 25.3 & 36.6 \\ \midrule
        AADA        & 92.2 & 87.3 & 88.3 & 90.0 & 59.9 & 90.2 & 69.7 & 46.1 & 44.0 & 45.7 
                    & 36.4 & 59.5 & 62.9 & 32.0 & 55.3 & 45.1 & 50.6 & 38.2 & 32.6 & 59.3 \\
        MADA        & 95.1 & 88.5 & 89.1 & 91.2 & 69.8 & 91.5 & 73.9 & 45.7 & 46.7 & 48.7 
                    & 43.3 & 59.2 & 68.7 & 48.4 & 60.6 & 56.9 & 53.3 & 50.1 & 51.6 & 64.9 \\
        RA (5\%)    & 97.0 & 90.4 & 90.2 & 92.7 & 77.3 & 93.2 & 75.0 & 47.7 & 59.2 & 53.2 
                    & 54.6 & 64.1 & 70.3 & 68.9 & 73.0 & 79.7 & 55.9 & 54.8 & 55.5 & 71.2 \\
        ILM-ASSL    & 96.9 & 91.6 & 91.9 & 94.9 & 77.8 & 94.5 & 82.3 & 63.2 & 54.9 & 56.0 
                    & 46.7 & 77.4 & 77.6 & 75.3 & 79.3 & 88.1 & 70.8 & 61.2 & 65.8 & 76.1 \\ \midrule 
        \midrule
        RA (5\%)                        & \textbf{97.5} & \textbf{90.9} & \textbf{91.0} &         93.3  & \textbf{79.9} & \textbf{94.0} &         74.5  &         52.5  &         59.7  &         53.0  
                                        &         54.7  &         66.2  &         71.1  &         61.6  &         79.1  &         78.7  & \textbf{57.4} &         53.5  &         52.8  &         71.7  \\
        \rowcolor{shadecolor}
        CBDA (5\%)                      &         97.3  &         90.7  &         90.7  & \textbf{93.4} &         79.4  &         93.7  &         74.8  & \textbf{53.3} &         60.4  &         53.2  
                                        & \textbf{56.6} & \textbf{66.5} & \textbf{71.3} & \textbf{69.8} &         77.8  & \textbf{80.1} &         55.6  &         53.0  &         55.9  & \textbf{72.3} \\
        \midrule
        RA (10\%)                       &         97.2  &         90.2  &         90.6  &         92.8  &         78.2  &         93.8  &         72.6  &         49.2  &         60.0  &         50.3  
                                        &         49.6  &         64.5  &         69.8  &         62.9  &         76.9  &         78.6  &         55.8  &         52.0  &         54.3  &         70.5  \\
        \rowcolor{shadecolor}
        CBDA (10\%)                     &         97.2  &         90.7  &         90.8  &         93.3  &         78.6  &         93.7  & \textbf{75.1} &         52.1  & \textbf{60.7} & \textbf{54.0} 
                                        &         54.4  &         65.8  &         71.0  &         64.7  & \textbf{79.5} &         79.8  &         56.3  & \textbf{56.2} & \textbf{58.8} &         72.2  \\
        \midrule
        RA (20\%)                       &         97.0  &         89.7  &         90.0  &         92.0  &         76.7  &         92.8  &         71.6  &         45.9  &         57.5  &         49.4  
                                        &         53.3  &         60.1  &         67.0  &         64.2  &         73.3  &         75.0  &         51.0  &         50.0  &         50.3  &         68.8  \\
        \rowcolor{shadecolor}
        CBDA (20\%)     & 97.4 & 90.6 & 90.6 & 93.0 & 79.6 & 93.9 & 74.2 & 50.5 & 60.0 & 52.2 & 53.5 & 65.2 & 70.3 & 62.1 & 77.2 & 79.3 & 54.9 & 54.5 & 53.8 & 71.2 \\ \midrule \midrule
        Fully Superv    & 97.4 & 91.1 & 91.1 & 93.6 & 77.9 & 93.2 & 74.7 & 51.9 & 57.8 & 53.7
                        & 54.9 & 64.7 & 71.3 & 67.8 & 76.4 & 79.3 & 57.9 & 54.8 & 55.6 & 71.8 \\ \bottomrule
    \end{tabular}
\end{table}

The results show that our novel Class Balanced Dynamic Acquisition method improves over the RA baseline for pixel-wise active domain adaptation.
The re-run of the 5\% budget RA experiment already showed an improved score of $71.7$ over the official one of $71.2$, this new method increased that to $72.3$ and so not only improved over the original performance by $0.6$ mIoU, but also exceeds the performance of the fully supervised approach given the same training setup.
This improved performance is more significant for the 10\% budget regime and even more for 20\%, with improvements of $1.7$ mIoU and $2.4$ mIoU, respectively.

Looking at the first few columns, the majority class performance remains stable for all CBDA methods and is even increased slightly for the class \textit{car} for example.
For the minority classes, there is consistent improvement for each of the CBDA methods over the RA experiments of the same budget.
This improvement is particularly high for the classes \textit{fence}, \textit{truck}, \textit{light} and \textit{rider}.
While for the majority class \textit{road}, the performance of RA with increasing budget drops from $97.5$ to $97.0$, some minority classes experience more significant drops and are therefore the primary causes for reduced overall performance.
The class IoU drops $6.6$ (\textit{pole}), $6.1$ (\textit{sign}) and $4.1$ (\textit{bike}) when using RA.
CBDA on the other hand manages to reduce the performance loss of those classes significantly, to only $2.8$, $1.3$, and $1.0$ respectively. \cref{sec:result_samples} shows some examples of samples, their ground truth labels and predictions using RA and CBDA for comparison.

\section{Concluding Remarks}

Using CBDA, the model is able to utilize the budget more effectively and showed minority class performance improvements while keeping stable performance for the majority classes.
It was applied to domain adaptive active learning here, but the method can be used without domain adaptation without change.
Furthermore, it is essentially heuristic agnostic, as long as matrices of pixel-wise scores are calculated.

While the performance loss for larger AL budgets has not been eliminated, using more balanced active labels both increased the base performance and slowed the degradation for larger budgets.
In addition, the performance exceeded the fully supervised scheme.
Previous works have stated that domain adaptive active learning approaches can likely beat a fully supervised model eventually \citep{LabOR}.
However, the fully supervised variant is still sometimes seen as the upper bound on performance \citet{RIPU}.
The experimental results here show domain-adaptive active learning outperforming a fully supervised domain-adaptive approach - at least with imbalanced ground truth data.

Currently, human annotating for semantic segmentation is not typically performed pixel wise and in a production setting the pixel-based approach would be more challenging than an image-based one.
However, this proposed method relies on the flexibility of selecting only specific pixels from an image.
Further development for real-world annotation tools would aide the adoption of active learning in a production setting while allowing pixel-based methods to be applied.

\vspace{-1ex}
\acks{
This work was performed during a master research internship at Stellantis.
The authors want to thank Ankit Singh for his support during this work.
}

\appendix
\renewcommand{\thesection}{\Alph{section}}
\renewcommand{\theHsection}{appendixsection.\Alph{section}}

\newpage
\section{ } %
\label{sec:da_usage}

\begin{figure}[H]
    \centering
    \includegraphics[width=0.9\textwidth]{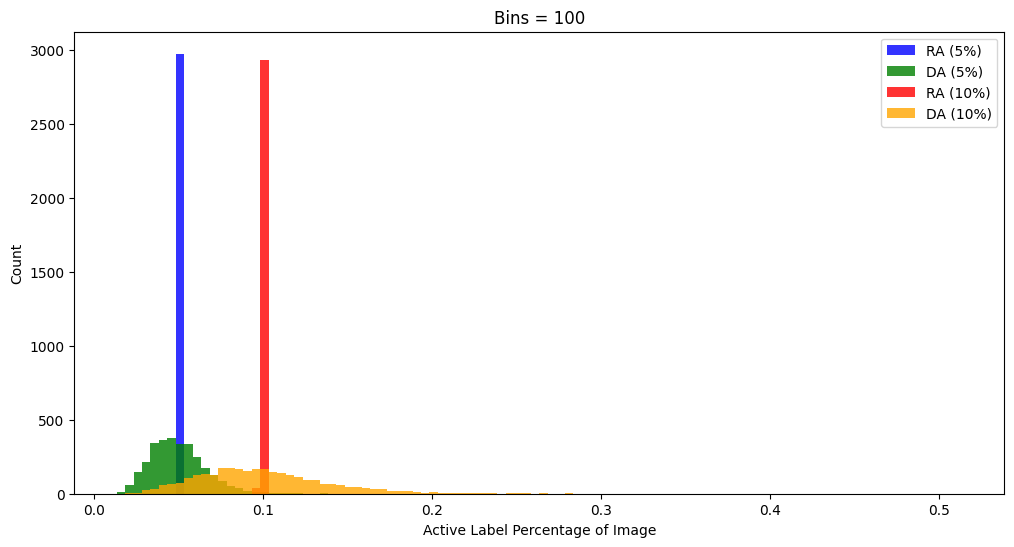}
    \caption{Histogram of selected pixel percentages showing broader spread of DA compared to RA}
    \label{fig:da_usage}
\end{figure}

\newpage
\section{  } %
\label{sec:cbda}

\begin{algorithm}
\setstretch{1.2}
\caption{Class Balanced Dynamic Acquisition (CB-DA)}
\label{alg:cbda}
\KwData{AL Budget $B_{AL}$, \# of AL Iterations $N_i$, \# of Classes $C$, Epsilon $\epsilon$, \\
Target Images $\{X_t\}$, AL Iteration $i$, Active Labels $\{\tilde{Y}_I\}$}
\vspace{5pt}
$matrices =$ list()\;
\;
\Comment{Statistics Gathering}
\ForEach{$(i,j)$ \textup{\textbf{in}} $\{\tilde{Y}_I\}$}
{
    $L_{c,i} \mathrel{+}= 1$, \textup{where} $\tilde{Y}_I^{(i,j)}$ \textup{==} c\;
}
\;
\Comment{Weight Calculation}
$B_{c,i} = (|\mathcal{T}| \times W \times H) \times \frac{B_{AL}}{C} \times \frac{i}{N_i}$\;
\ForEach{$c$ \textup{\textbf{in}} $\{1, \ldots, C\}$}
{
    $W_{c,i} = \max \left(1 - \frac{L_{c,i}}{B_{c,i}}, \epsilon \right)$\;
}
\;
\Comment{Weighted Scoring}
\ForEach{$I$ \textup{\textbf{in}} $\{X_t\}$}
{
    $\mathcal{A} = calculate\_scores(I)$\;
    \ForEach{$(i,j)$ \textup{\textbf{in}} $\mathcal{A}$}
    {
        $\hat{\mathcal{A}}_c^{(i,j)} = \mathcal{A}_c^{(i,j)} \times W_{c,i},$ \textup{where} $\hat{Y}_I^{(i,j)}$ \textup{==} c\;
    }
    $matrices$.$append(\hat{\mathcal{A}})$\;
}
\;
$\mathcal{A}_S = matrices$.$stack()$\;
$selection = select\_pixels(\mathcal{A}_S, iteration\_budget)$\;
\;
\ForEach{$(I,i,j)$ \textup{\textbf{in}} $selection$}
{
    $\tilde{Y}_I^{(i,j)} = Y_I^{(i,j)}$\;
}
\end{algorithm}

\newpage
\section{   } %
\label{sec:results_synthia}

\begin{table}[ht]
    \scriptsize
    \centering
    \caption{Results of the SYNTHIA $\rightarrow$ Cityscapes task}
    \label{tab:results_synthia}
    \setlength{\tabcolsep}{2pt}
    \begin{tabular}
    {
        @{}
        lccccccccccccccccc
        @{}
    }
        \toprule
        Method                          
        & \rotatebox{60}{road} 
        & \rotatebox{60}{buil.}
        & \rotatebox{60}{veg.}
        & \rotatebox{60}{car}
        & \rotatebox{60}{side.}
        & \rotatebox{60}{sky}
        & \rotatebox{60}{pers.}
        & \rotatebox{60}{pole}
        & \rotatebox{60}{fence}
        & \rotatebox{60}{wall}
        & \rotatebox{60}{sign}
        & \rotatebox{60}{bike}
        & \rotatebox{60}{bus}
        & \rotatebox{60}{light}
        & \rotatebox{60}{rider}
        & \rotatebox{60}{motor.}
        & mIoU \\ \midrule
        Source only     & 64.3 & 73.1 & 63.1 & 73.1 & 21.3 & 67.6 & 42.2 & 31.4 &  1.1 
                        &  2.4 & 27.7 & 38.9 & 15.3 &  7.0 & 19.9 & 10.5 & 34.9 \\ \midrule
        AADA (5\%)      & 91.3 & 86.9 & 88.2 & 89.9 & 57.6 & 90.3 & 69.4 & 45.0 & 48.3 
                        & 37.6 & 58.5 & 62.5 & 44.5 & 50.4 & 37.9 & 32.8 & 61.9 \\
        MADA (5\%)      & 96.5 & 88.8 & 89.7 & 90.9 & 74.6 & 92.2 & 74.1 & 46.7 & 43.8 
                        & 45.9 & 60.5 & 69.4 & 60.3 & 52.4 & 51.2 & 52.4 & 68.1 \\
        RA (5\%)        & 97.0 & 89.9 & 91.1 & 92.9 & 78.9 & 93.0 & 74.4 & 48.5 & 50.7 
                        & 47.2 & 63.9 & 71.0 & 79.9 & 55.2 & 54.1 & 55.3 & 71.4 \\
        ILM-ASSL (5\%)  & 97.4 & 91.8 & 91.6 & 94.4 & 80.1 & 94.5 & 82.7 & 64.1 & 55.2 
                        & 38.6 & 78.7 & 77.2 & 81.7 & 70.9 & 60.1 & 66.8 & 76.6 \\ \midrule \midrule
        RA (5\%)                        & \textbf{97.3} &         89.9  &         91.3  &         92.9  & \textbf{80.2} &         93.5  &         73.2  &         46.7  
                                        &         50.1  &         45.6  &         63.2  &         69.8  &         78.2  &         51.4  &         49.6  &         52.5  &         70.3  \\ \rowcolor{shadecolor}
        CBDA (5\%)                      &         97.1  &         89.9  &         90.9  &         92.9  &         79.1  &         93.7  &         73.7  &         46.1  
                                        &         50.7  &         46.0  &         63.4  &         70.3  &         78.8  &         51.2  &         51.4  &         47.5  &         70.2  \\ \midrule
        RA (10\%)                       & \textbf{97.3} &         90.2  & \textbf{91.5} &         93.0  &         80.0  &         93.8  &         73.7  & \textbf{49.4} 
                                        &         50.1  &         48.2  &         64.8  &         70.7  &         78.0  &         52.9  &         51.5  &         55.6  &         71.3  \\ \rowcolor{shadecolor}
        CBDA (10\%)                     &         97.2  & \textbf{90.4} &         91.2  & \textbf{93.3} &         79.7  &         93.7  &         74.1  &         48.6  
                                        & \textbf{51.1} & \textbf{52.7} &         65.3  &         71.2  & \textbf{82.0} & \textbf{53.4} & \textbf{53.7} &         56.1  & \textbf{72.1} \\ \midrule 
        RA (20\%)                       & \textbf{97.3} &         90.2  & \textbf{91.5} &         93.2  & \textbf{80.2} & \textbf{93.9} & \textbf{74.3} &         48.0  
                                        &         46.6  &         47.2  &         64.5  &         70.7  &         73.8  &         52.8  &         52.4  &         55.0  &         70.7  \\ \rowcolor{shadecolor}
        CBDA (20\%)                     &         97.1  &         89.6  &         90.8  &         93.0  &         79.5  &         93.6  &         74.0  &         45.2  
                                        &         50.4  &         50.3  & \textbf{66.1} & \textbf{71.3} &         77.4  &         49.8  &         53.2  & \textbf{57.8} &         71.2  \\ \midrule \midrule
        Fully Superv.   & 97.5 & 90.9 & 91.7 & 93.2 & 81.4 & 93.4 & 75.6 & 53.6 & 51.3 & 48.5 & 68.1 & 71.2 & 75.6 & 59.4 & 51.9 & 52.0 & 72.2 \\ \bottomrule
    \end{tabular}
\end{table}

\newpage
\section{    } %
\label{sec:result_samples}

\begin{table}[ht]
    \centering
    \caption{Example results for RA and CBDA for 5\% and 10\%}
    \label{tab:results_images}
    \setlength{\tabcolsep}{2pt}
    \begin{tabular}{ccc}
        Sample & RA (5\%) & RA (10\%) \\
        \includegraphics[width=0.32\textwidth]{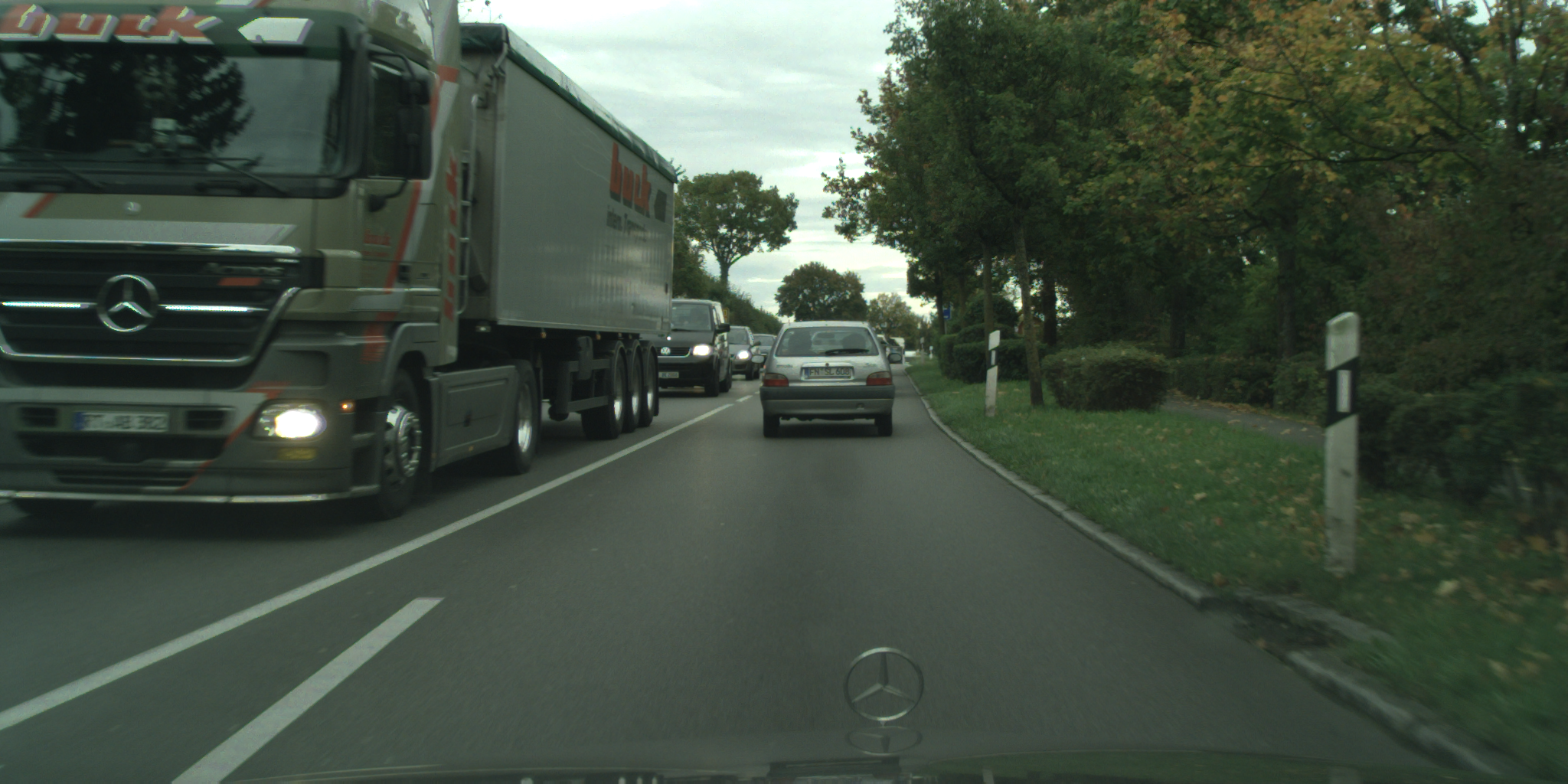} & 
        \includegraphics[width=0.32\textwidth]{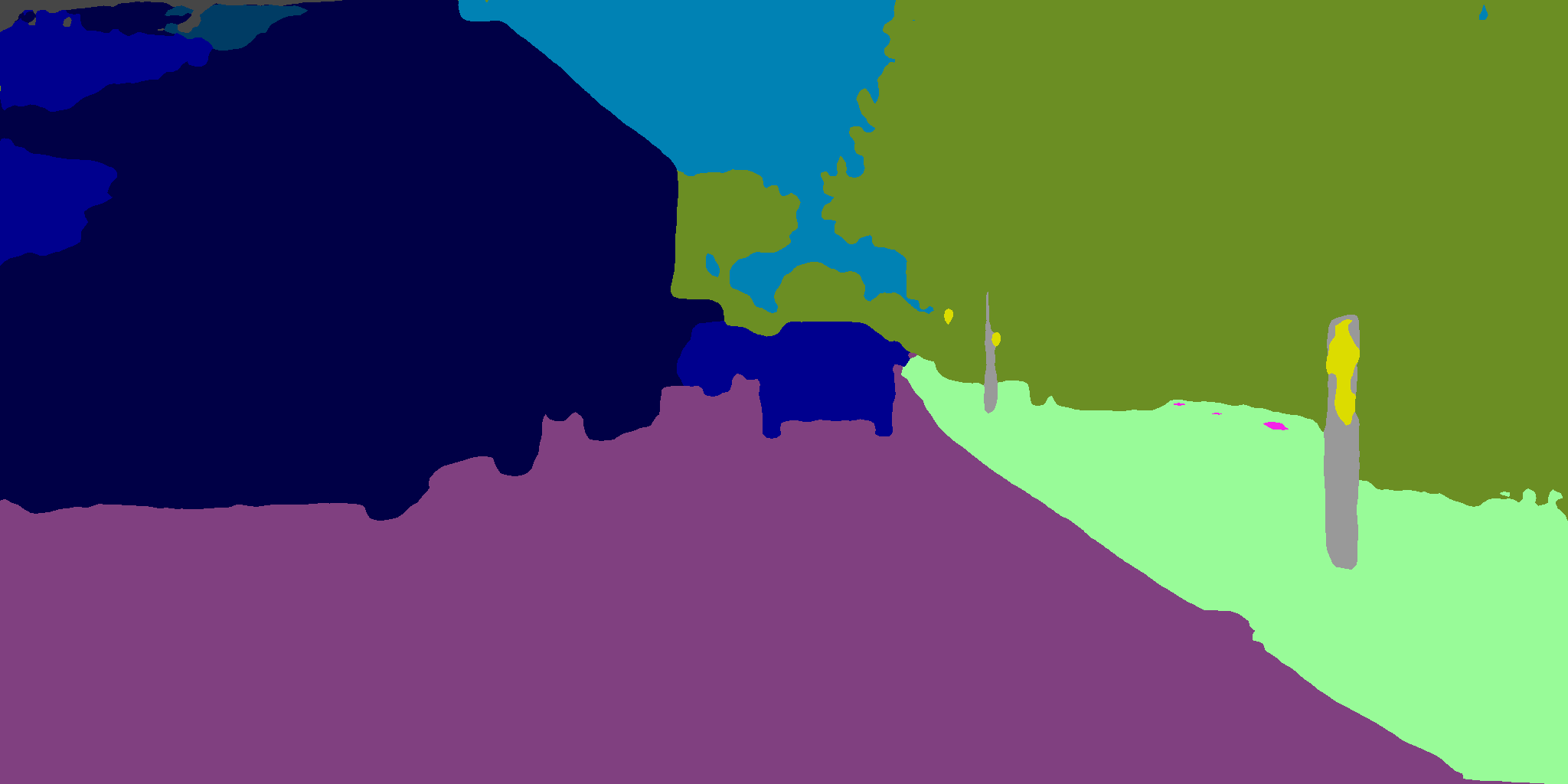} &
        \includegraphics[width=0.32\textwidth]{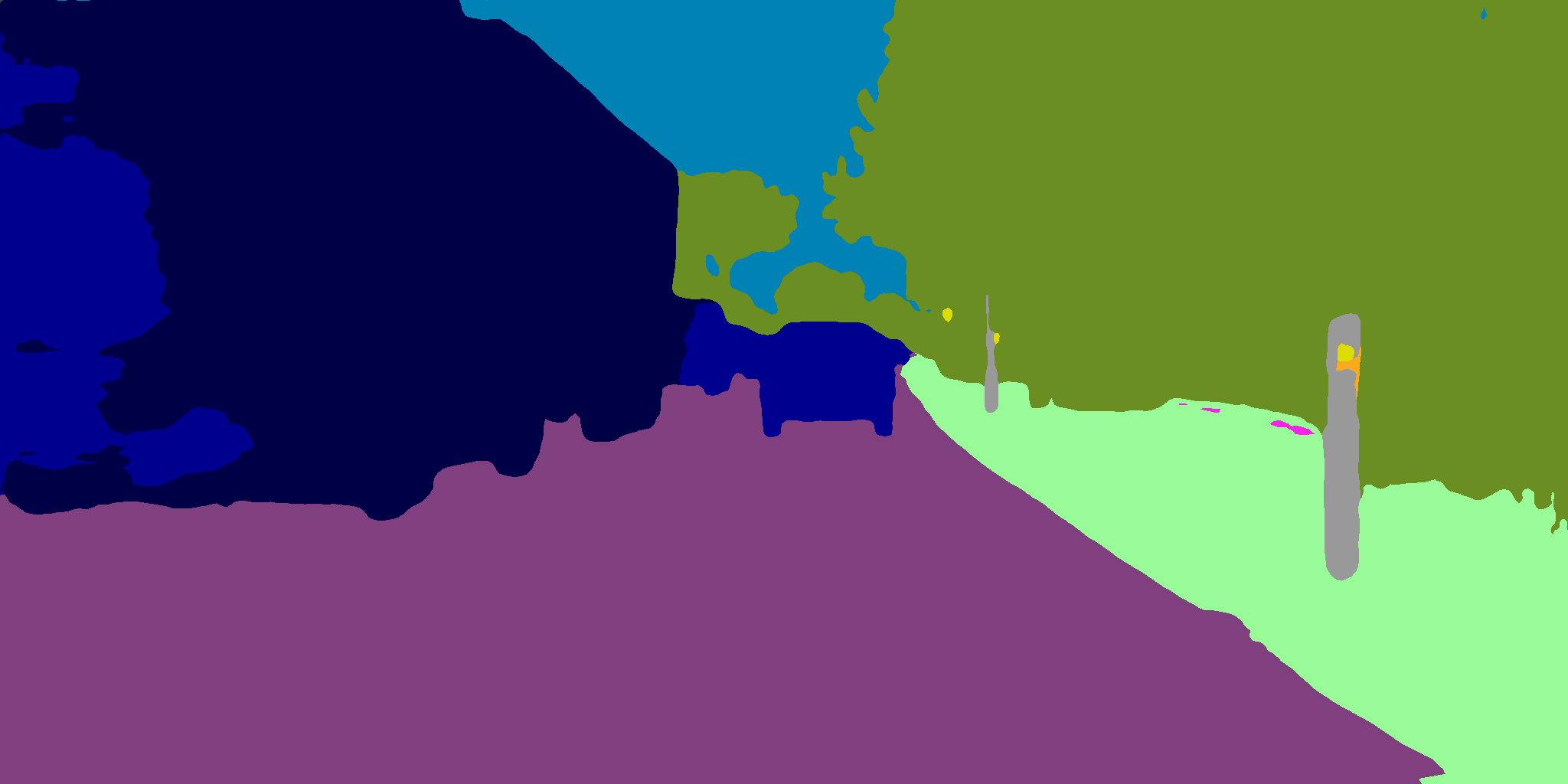} \\
        Label & CBDA (5\%) & CBDA (5\%) \\
        \includegraphics[width=0.32\textwidth]{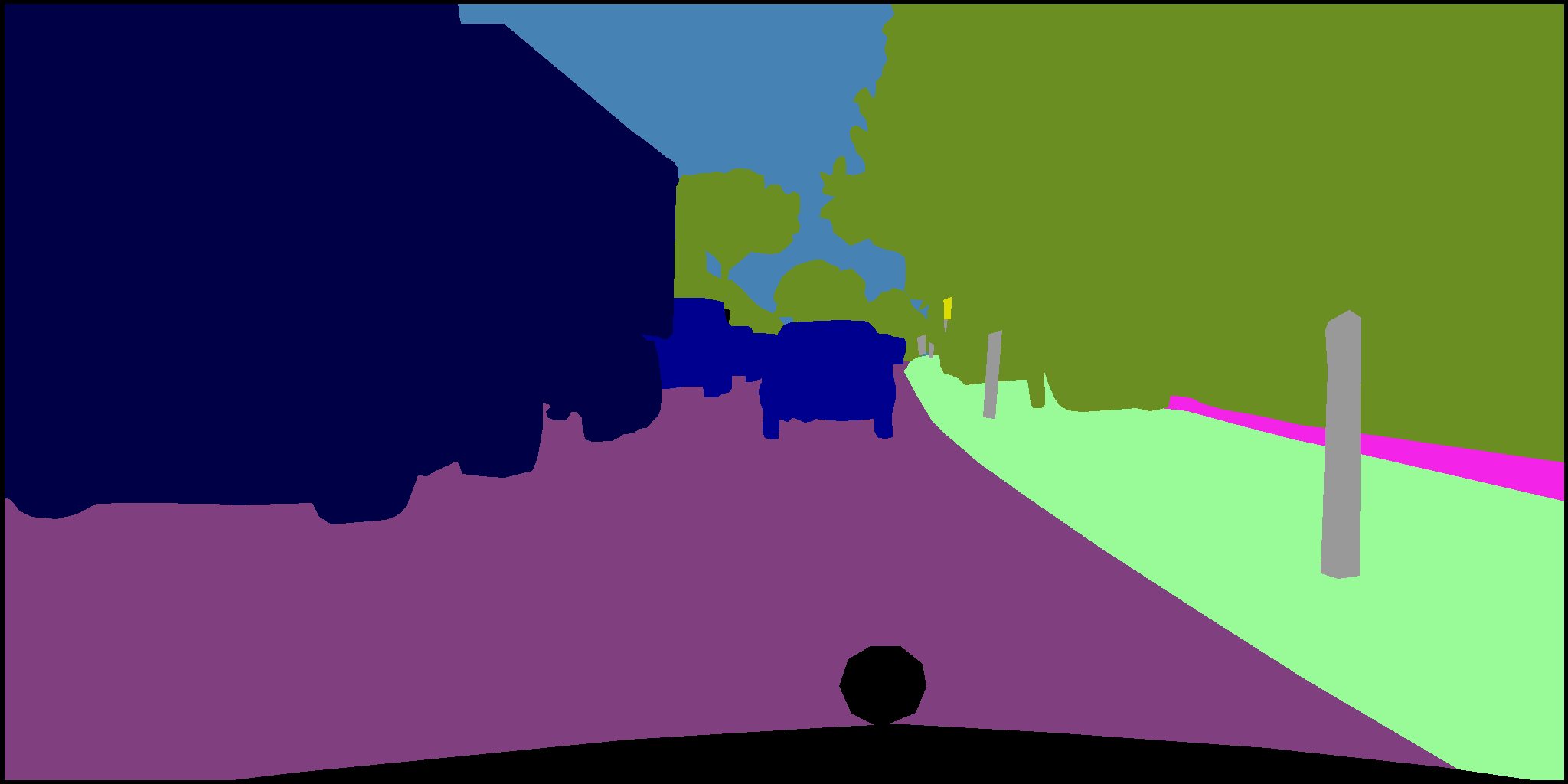} & 
        \includegraphics[width=0.32\textwidth]{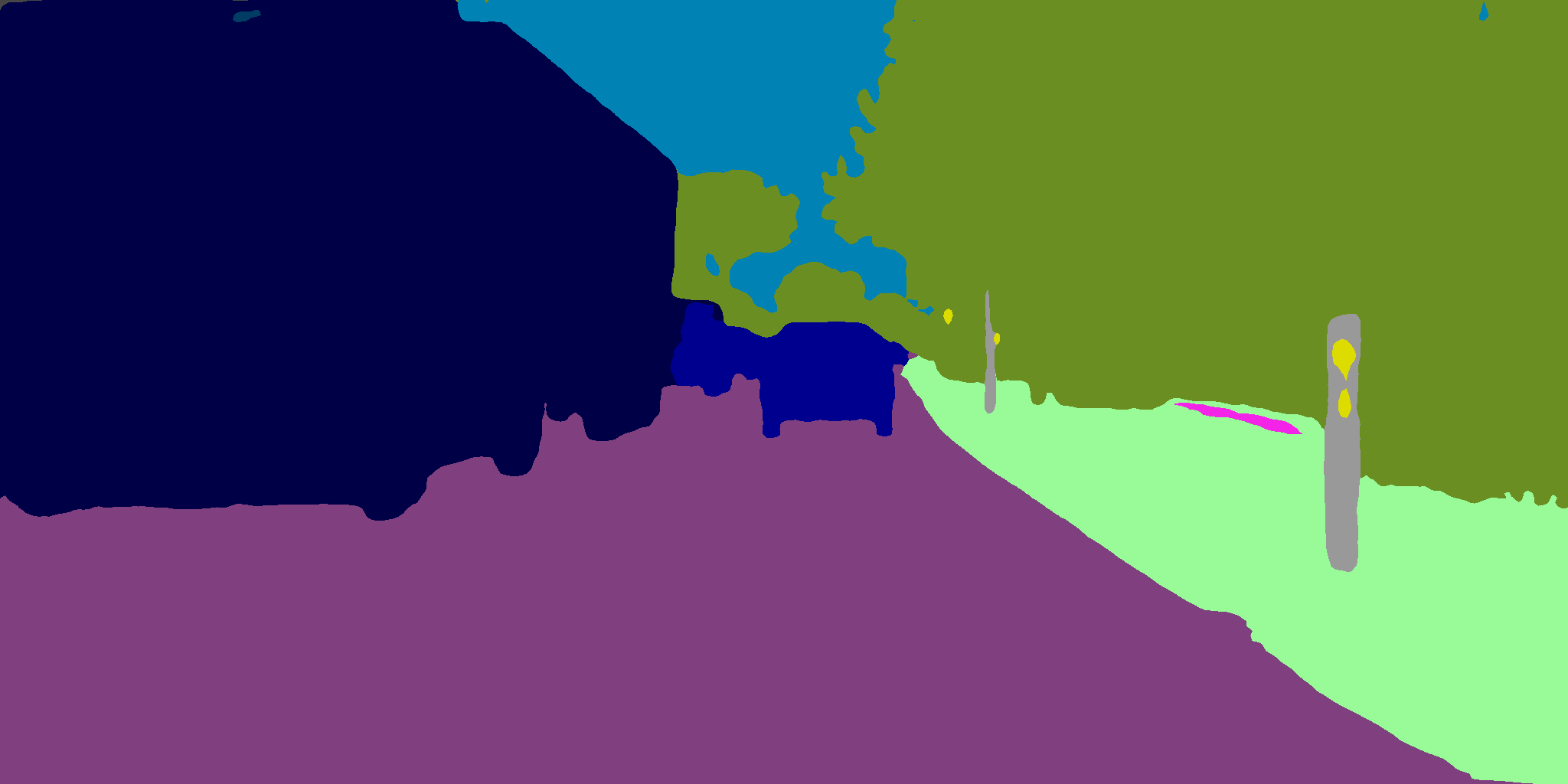} &
        \includegraphics[width=0.32\textwidth]{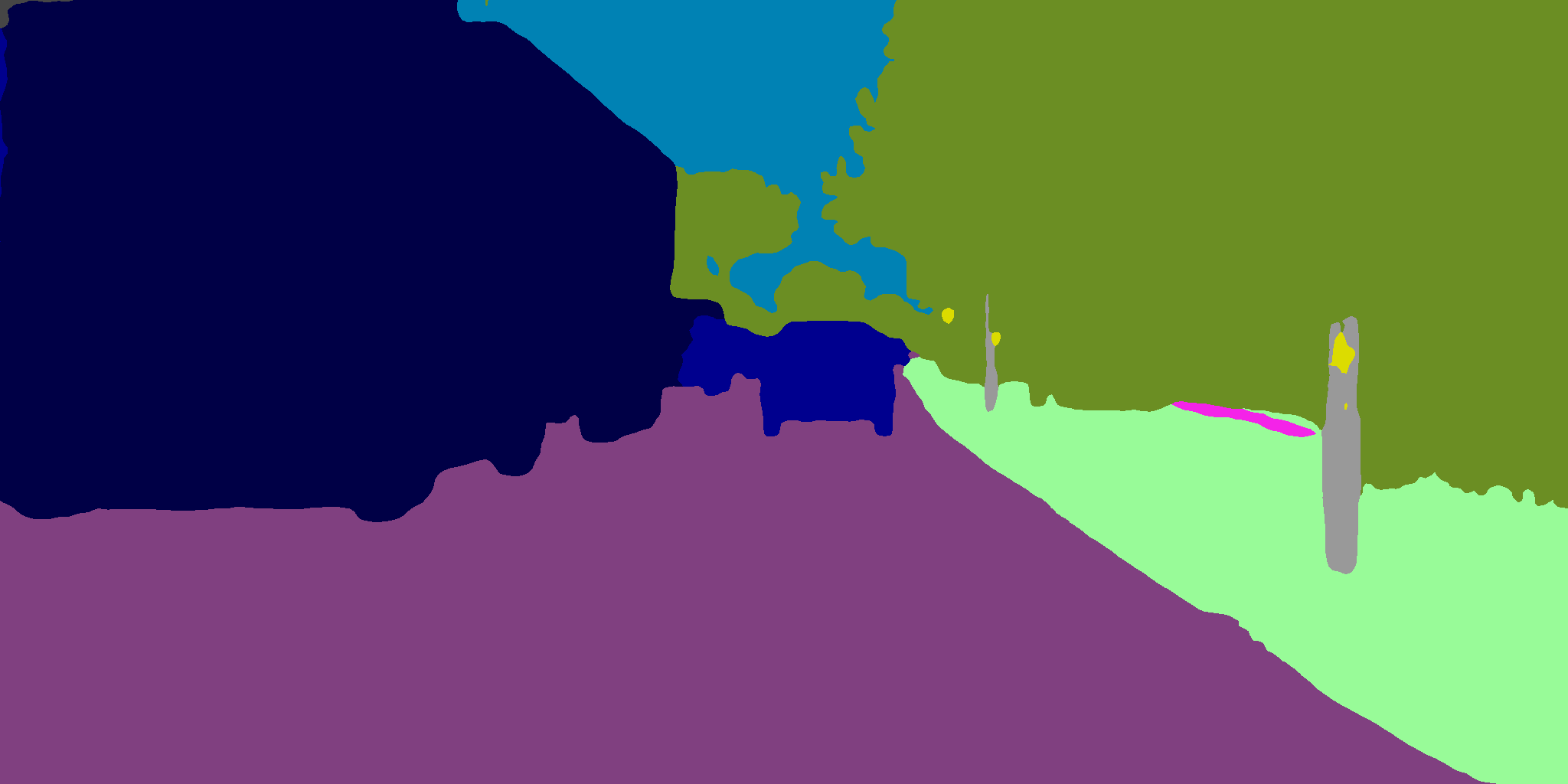} \\
        \\
        \includegraphics[width=0.32\textwidth]{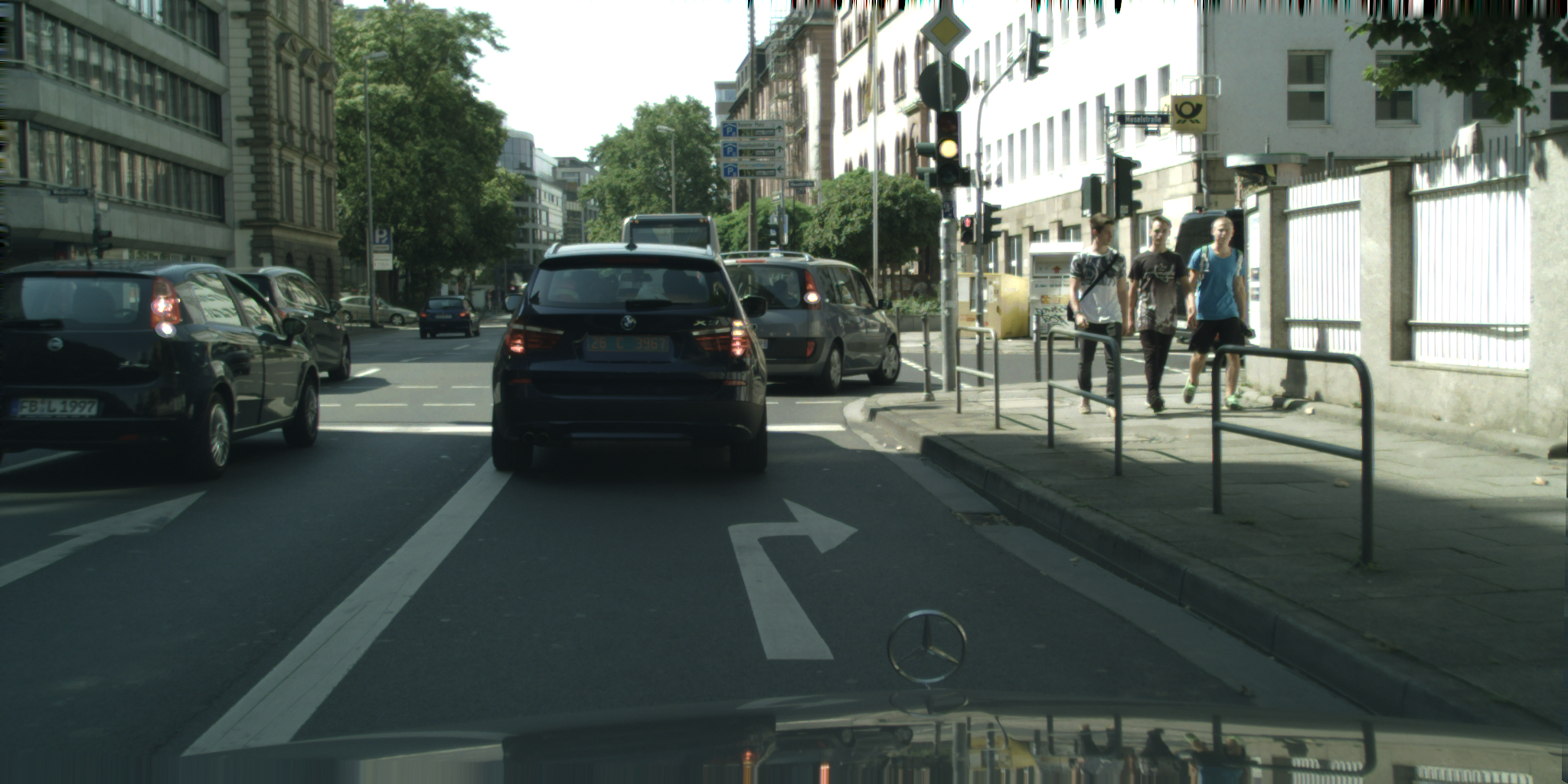} & 
        \includegraphics[width=0.32\textwidth]{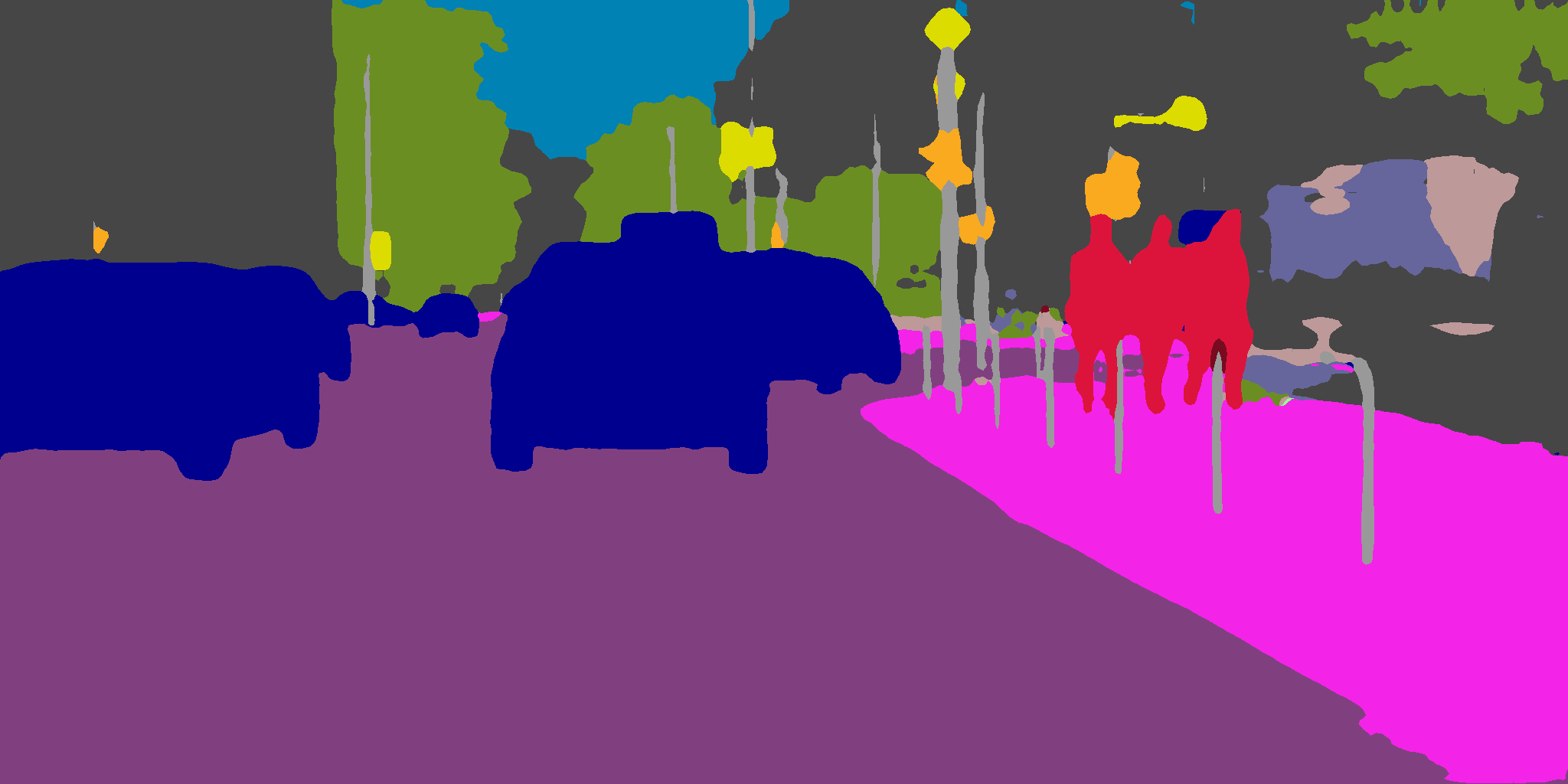} &
        \includegraphics[width=0.32\textwidth]{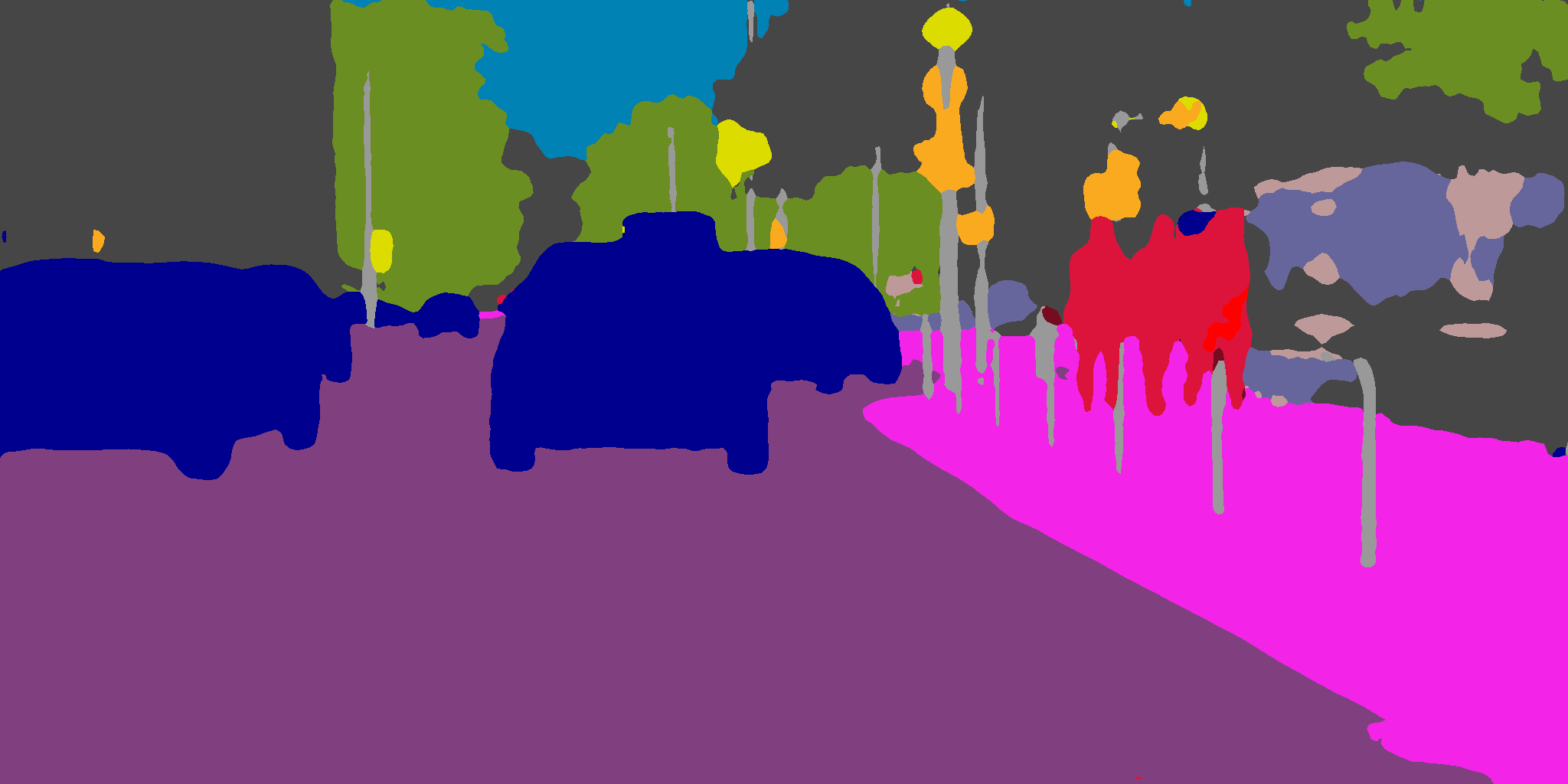} \\
        \includegraphics[width=0.32\textwidth]{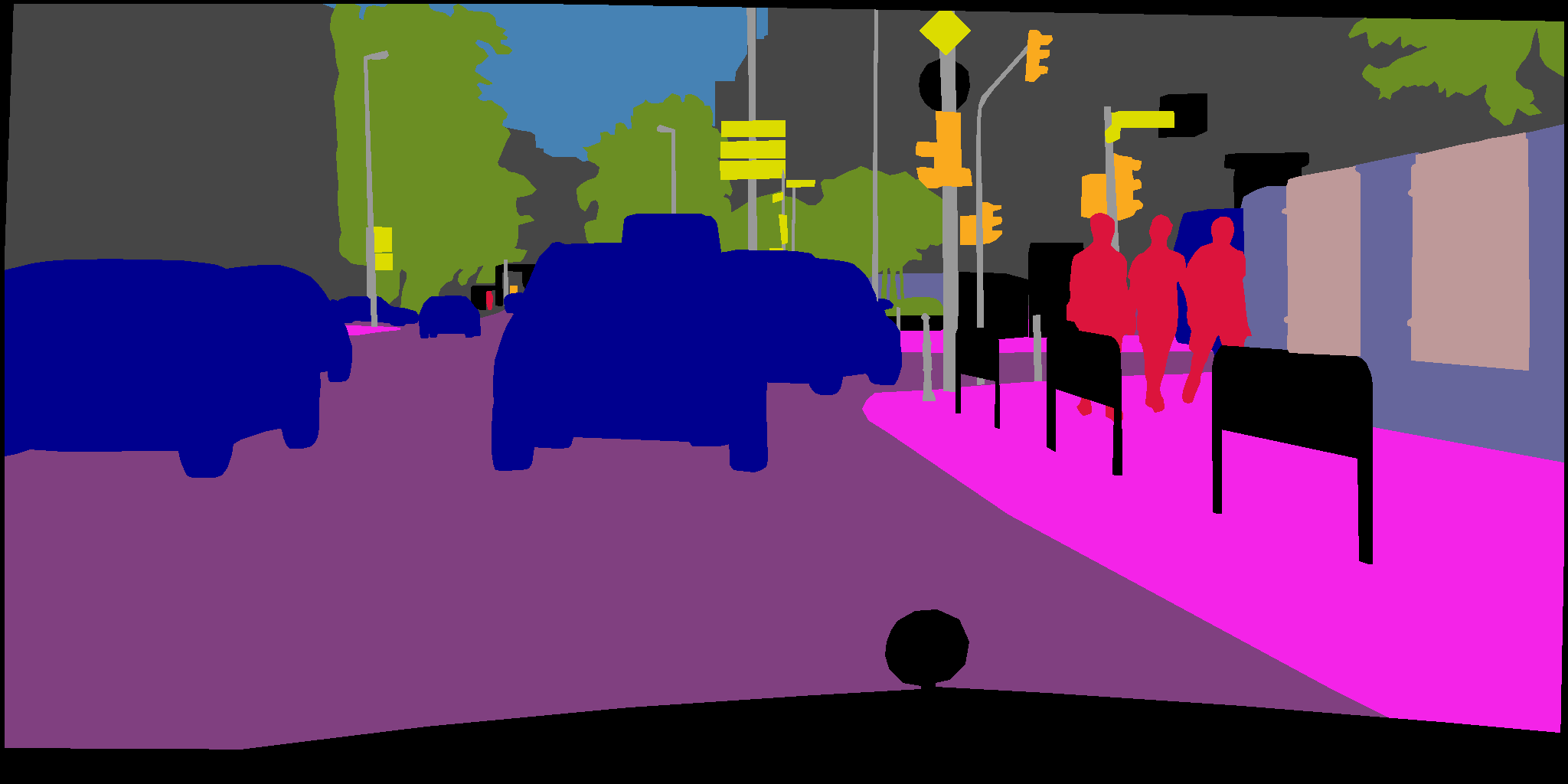} & 
        \includegraphics[width=0.32\textwidth]{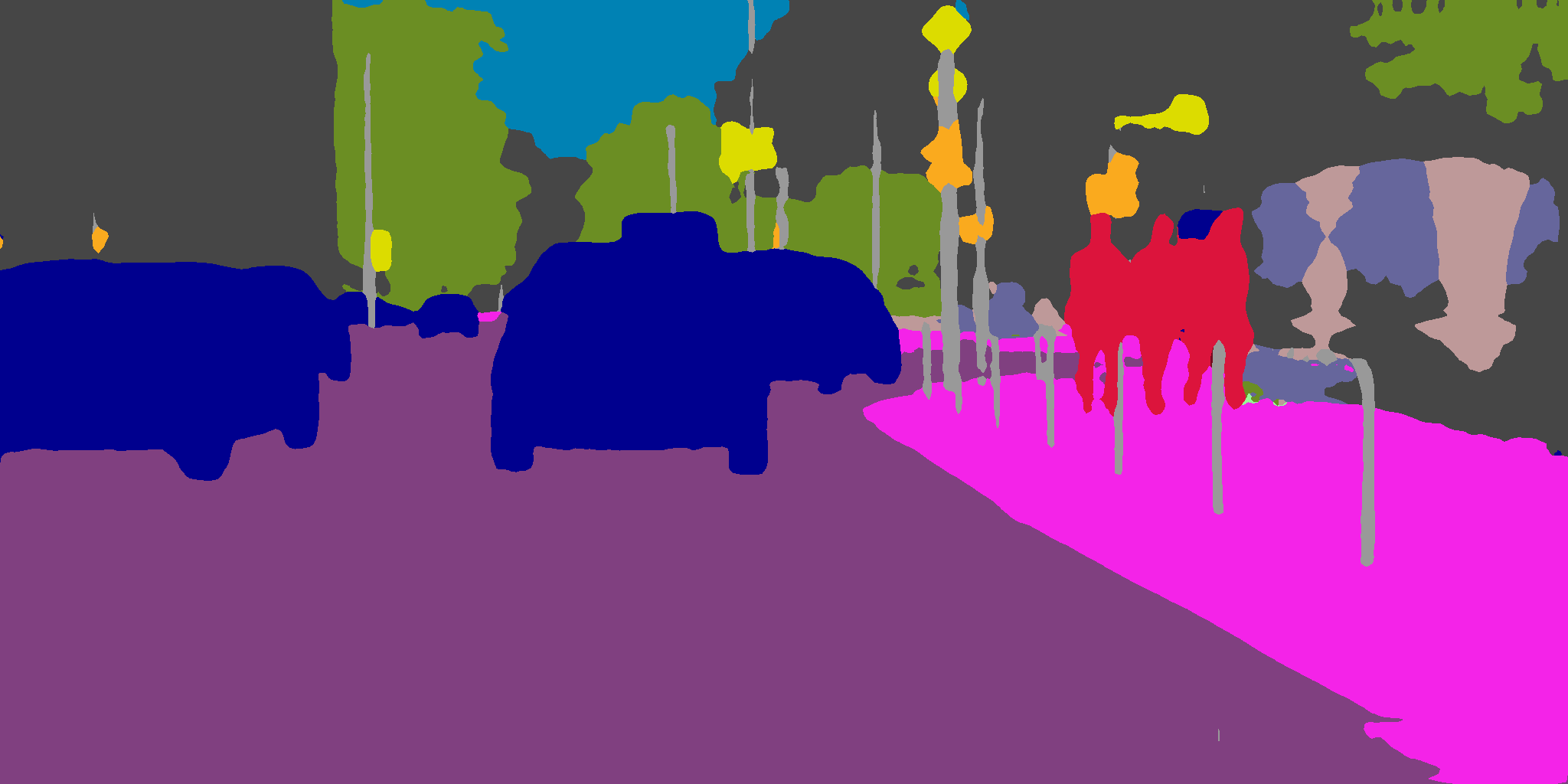} &
        \includegraphics[width=0.32\textwidth]{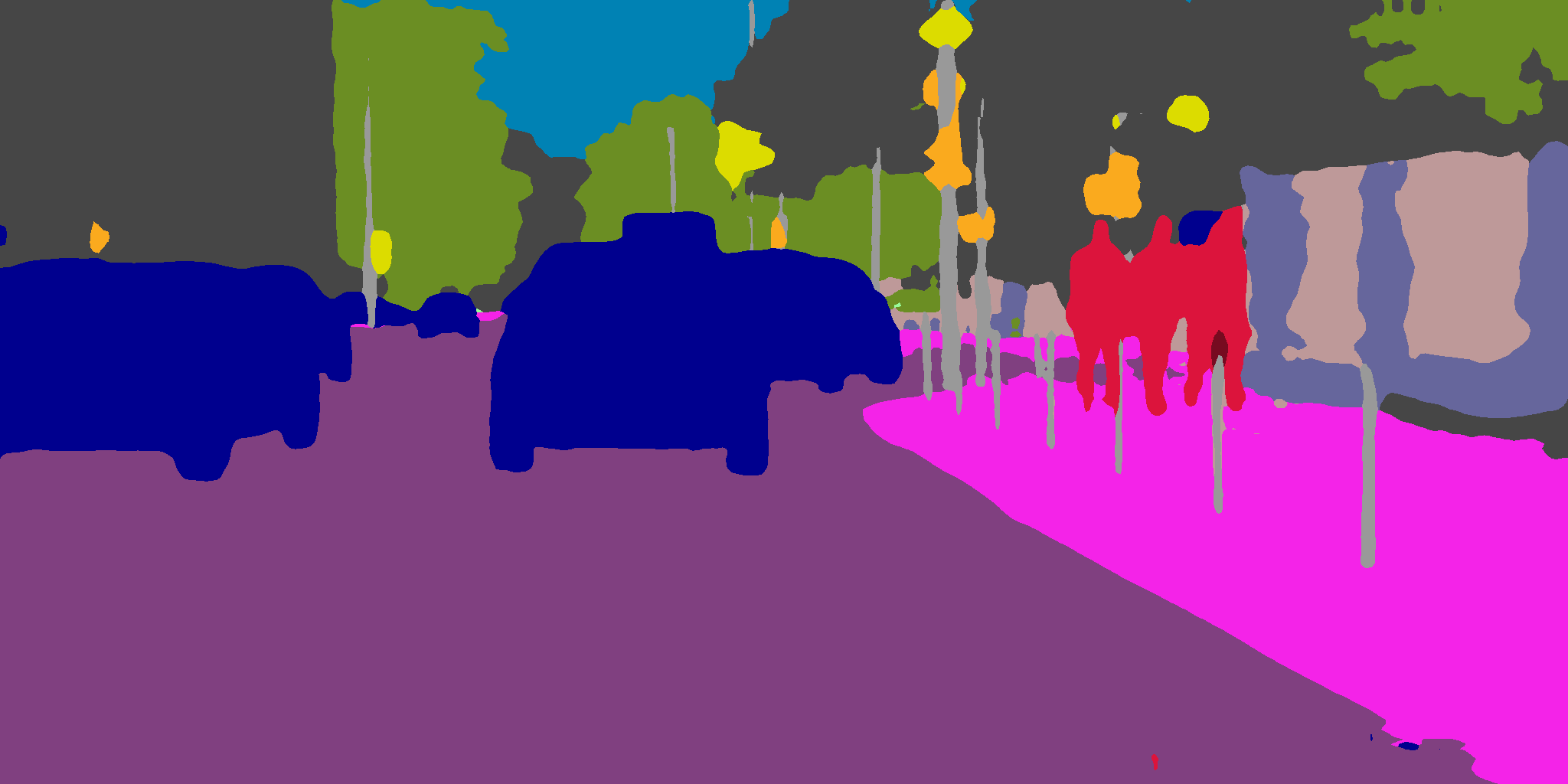} \\
        \\
        \includegraphics[width=0.32\textwidth]{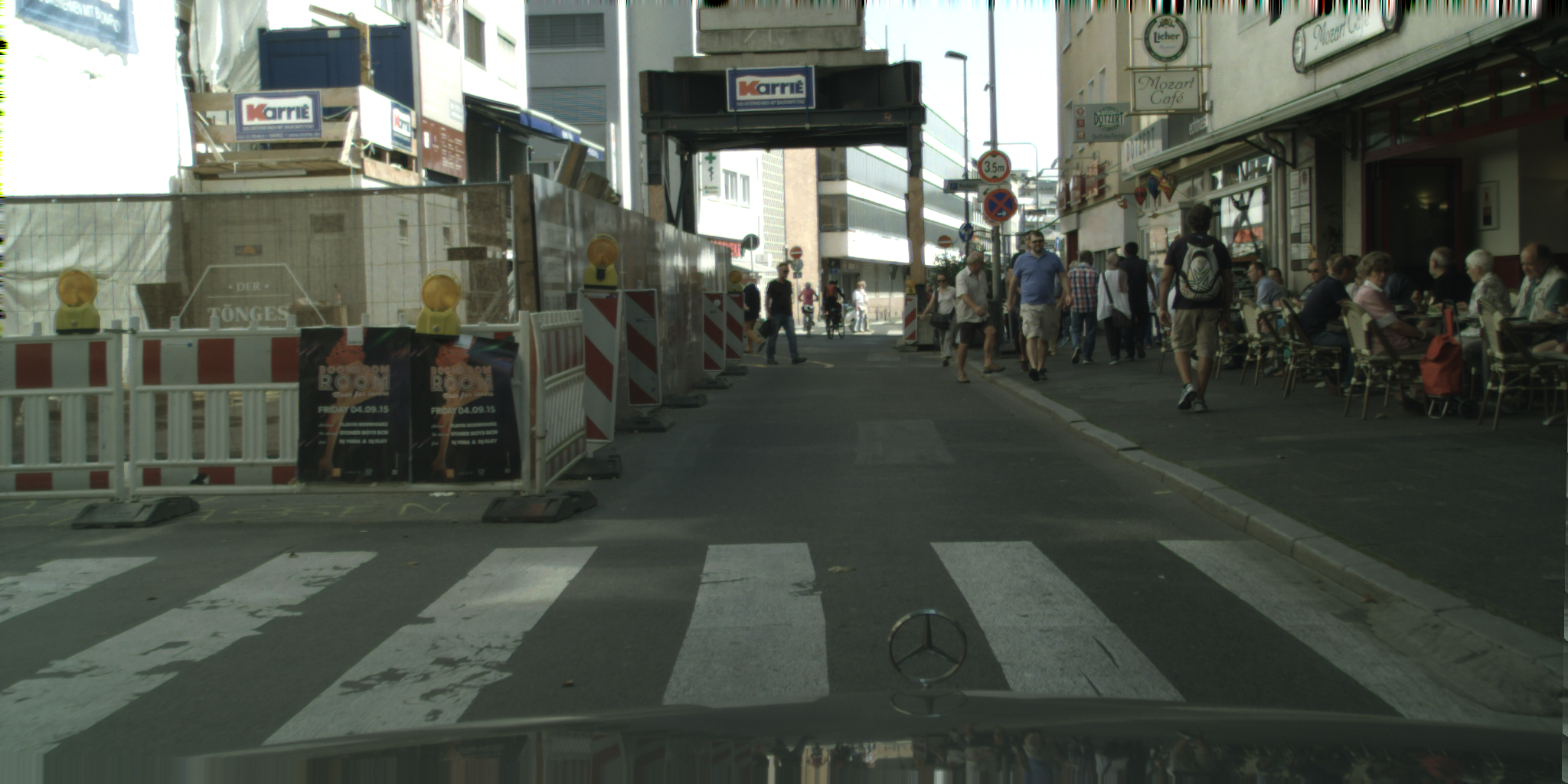} & 
        \includegraphics[width=0.32\textwidth]{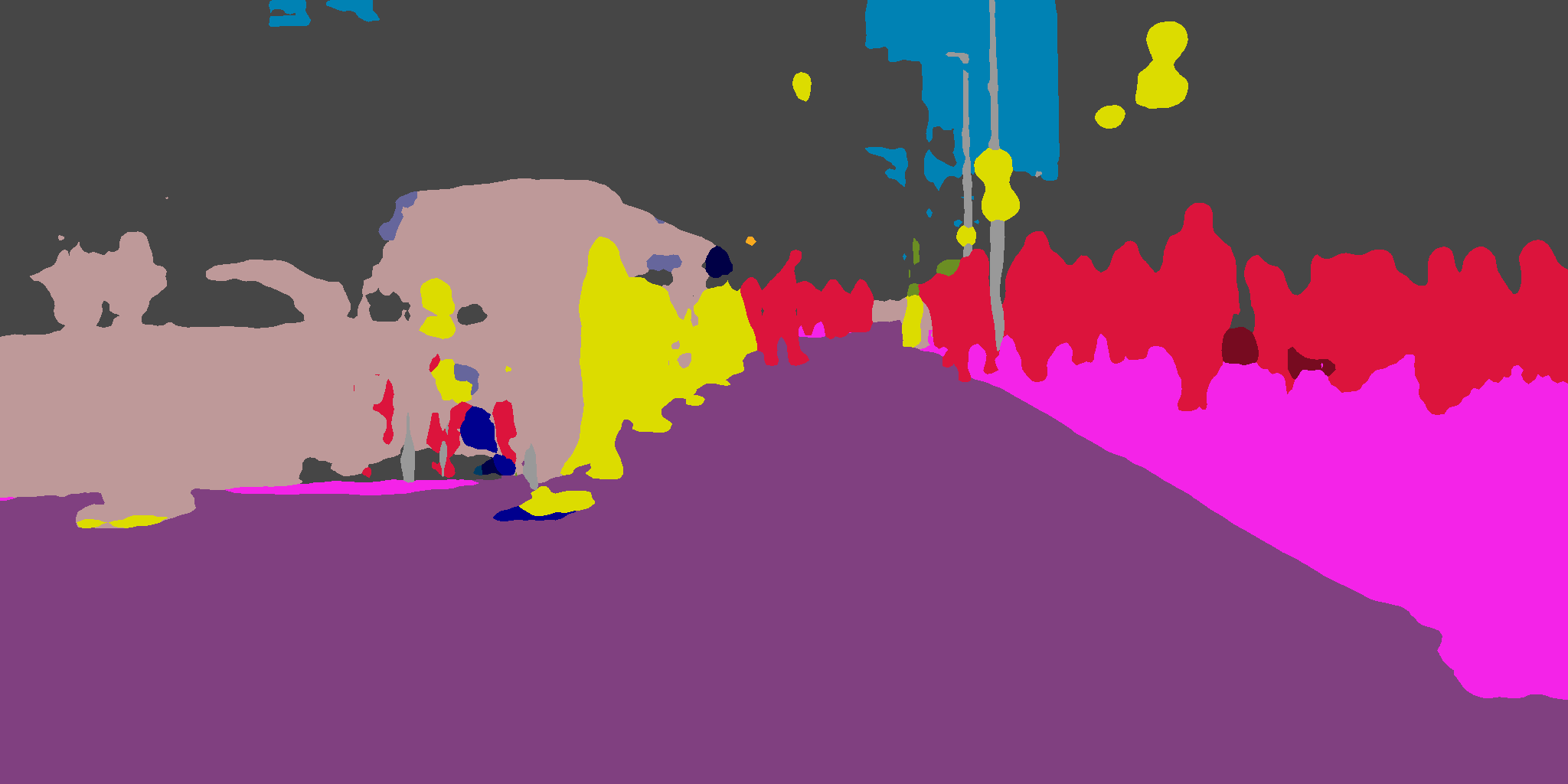} &
        \includegraphics[width=0.32\textwidth]{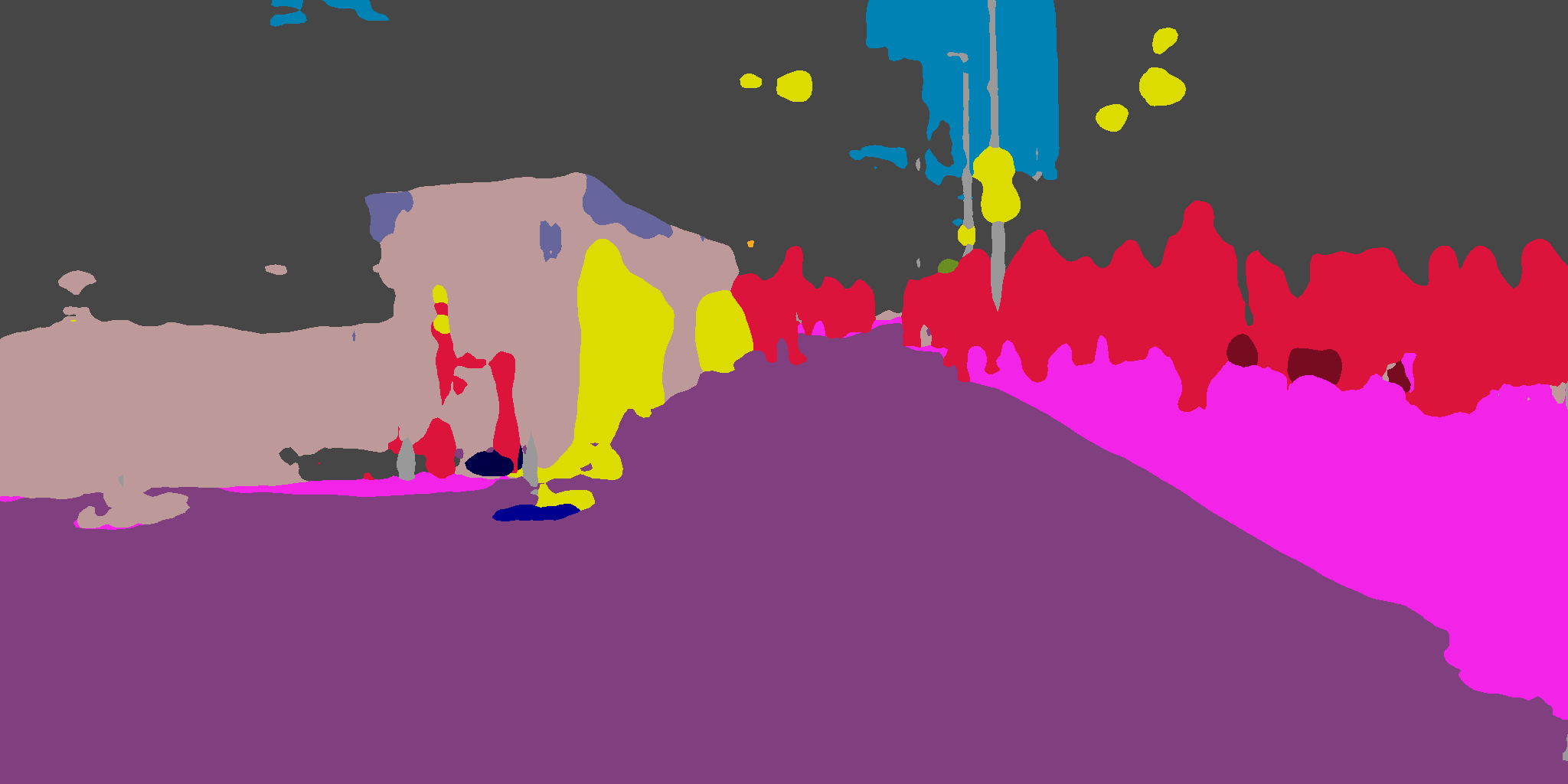} \\
        \includegraphics[width=0.32\textwidth]{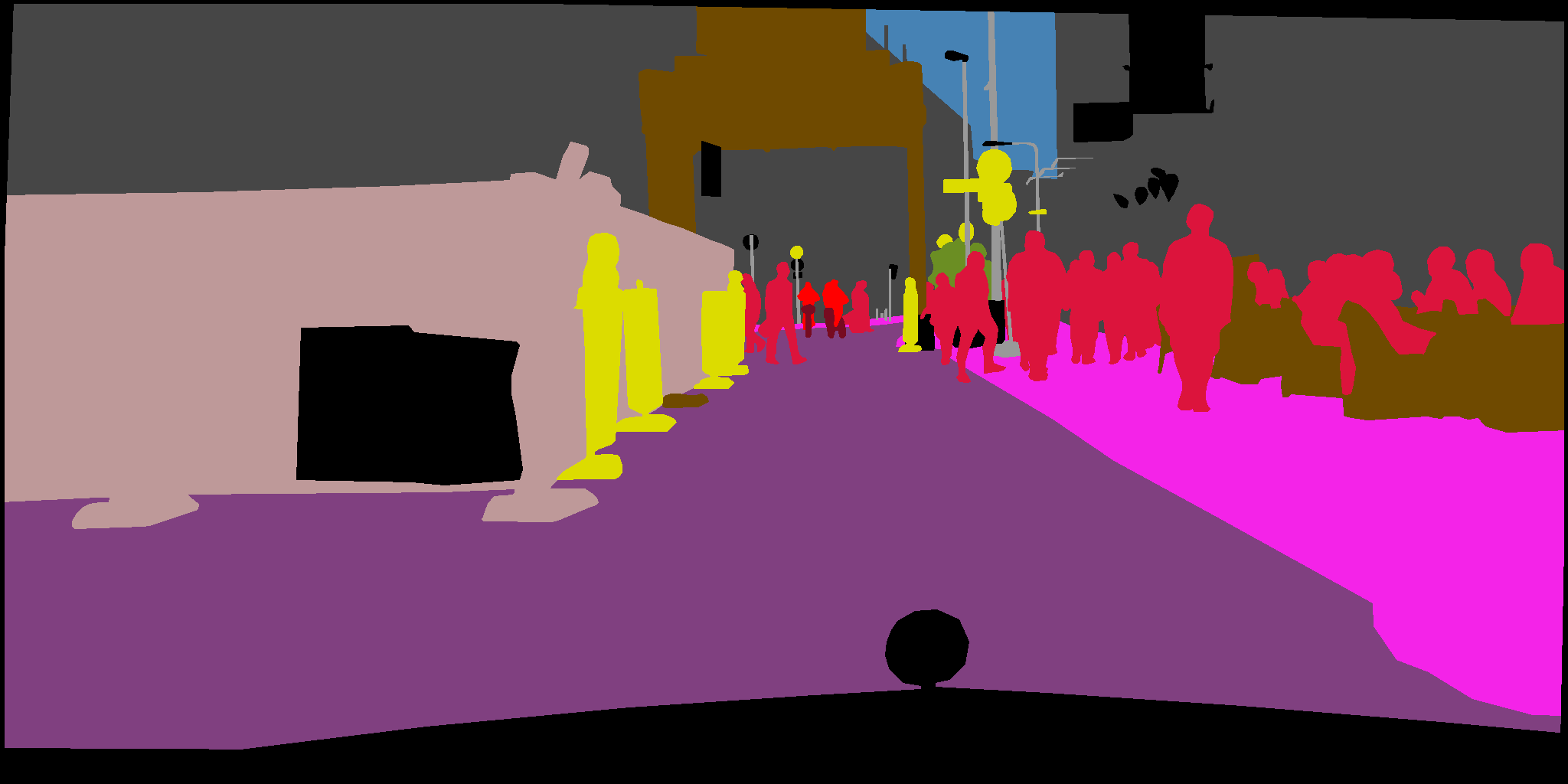} & 
        \includegraphics[width=0.32\textwidth]{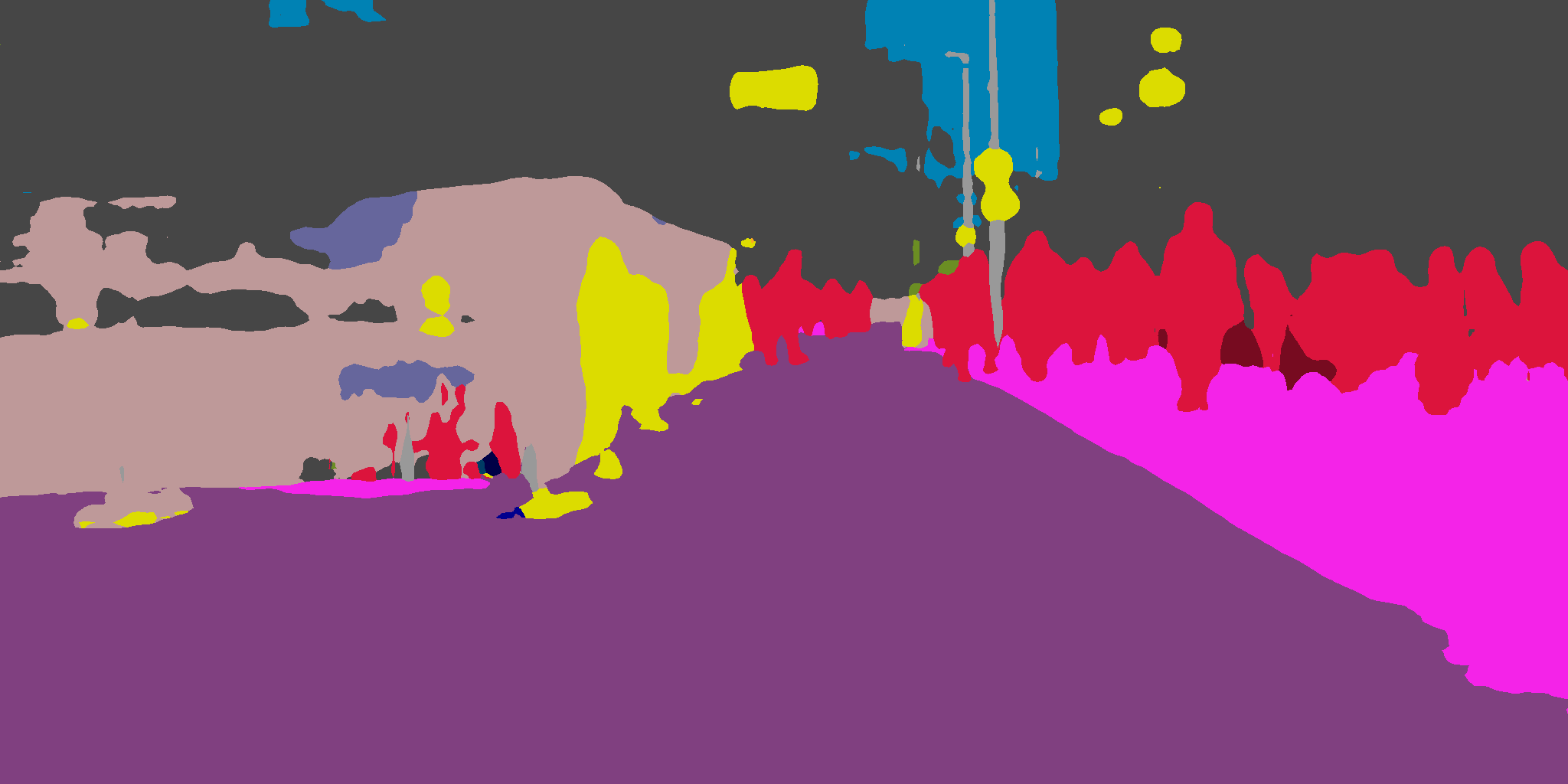} &
        \includegraphics[width=0.32\textwidth]{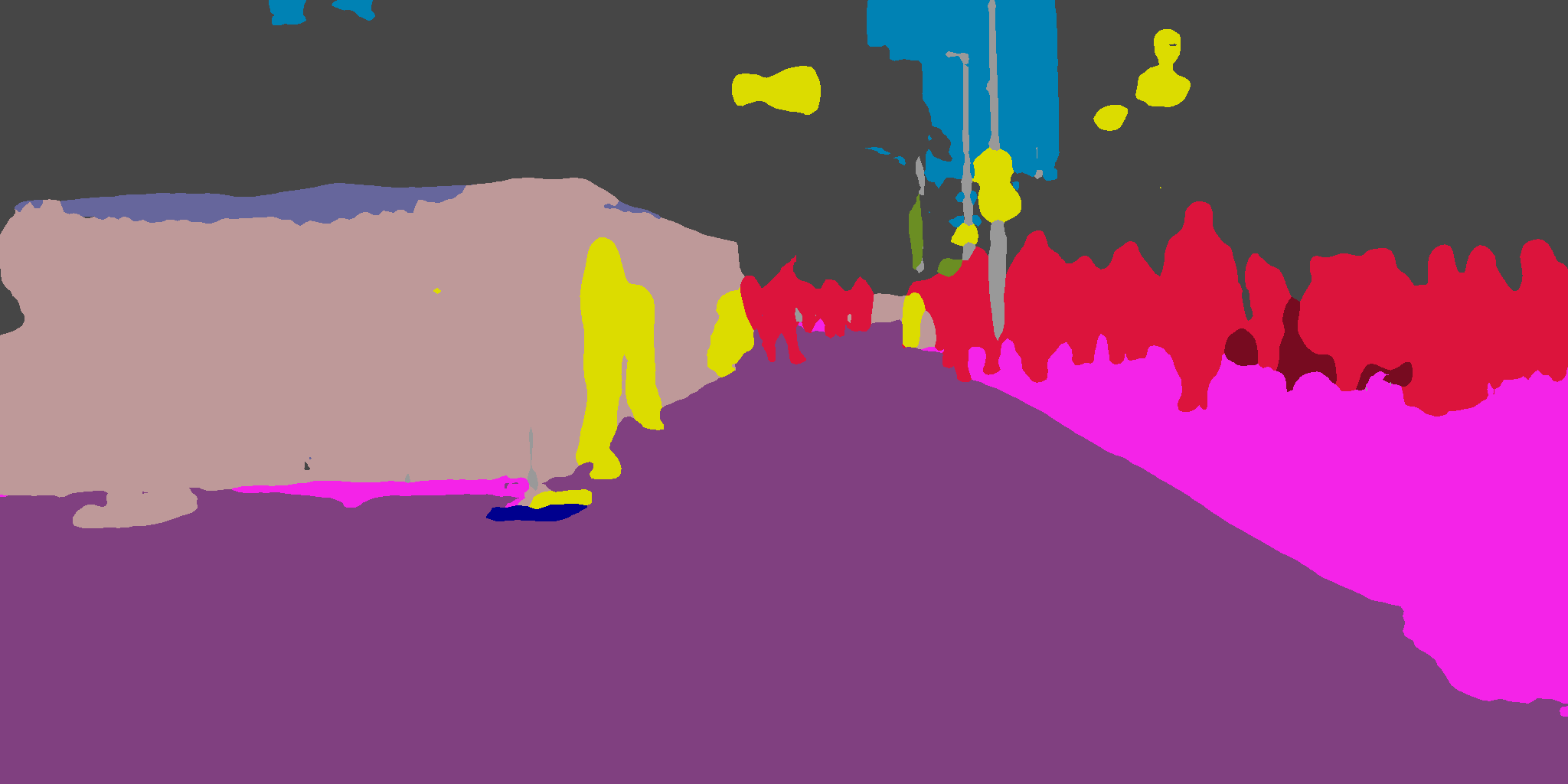} \\
    \end{tabular}
\end{table}

\newpage
\section{     } %
\label{sec:ablation_minority}

\subsection*{Minority Class Improvements}

\begin{figure}[ht]
    \centering
    \includegraphics[width=0.95\textwidth]{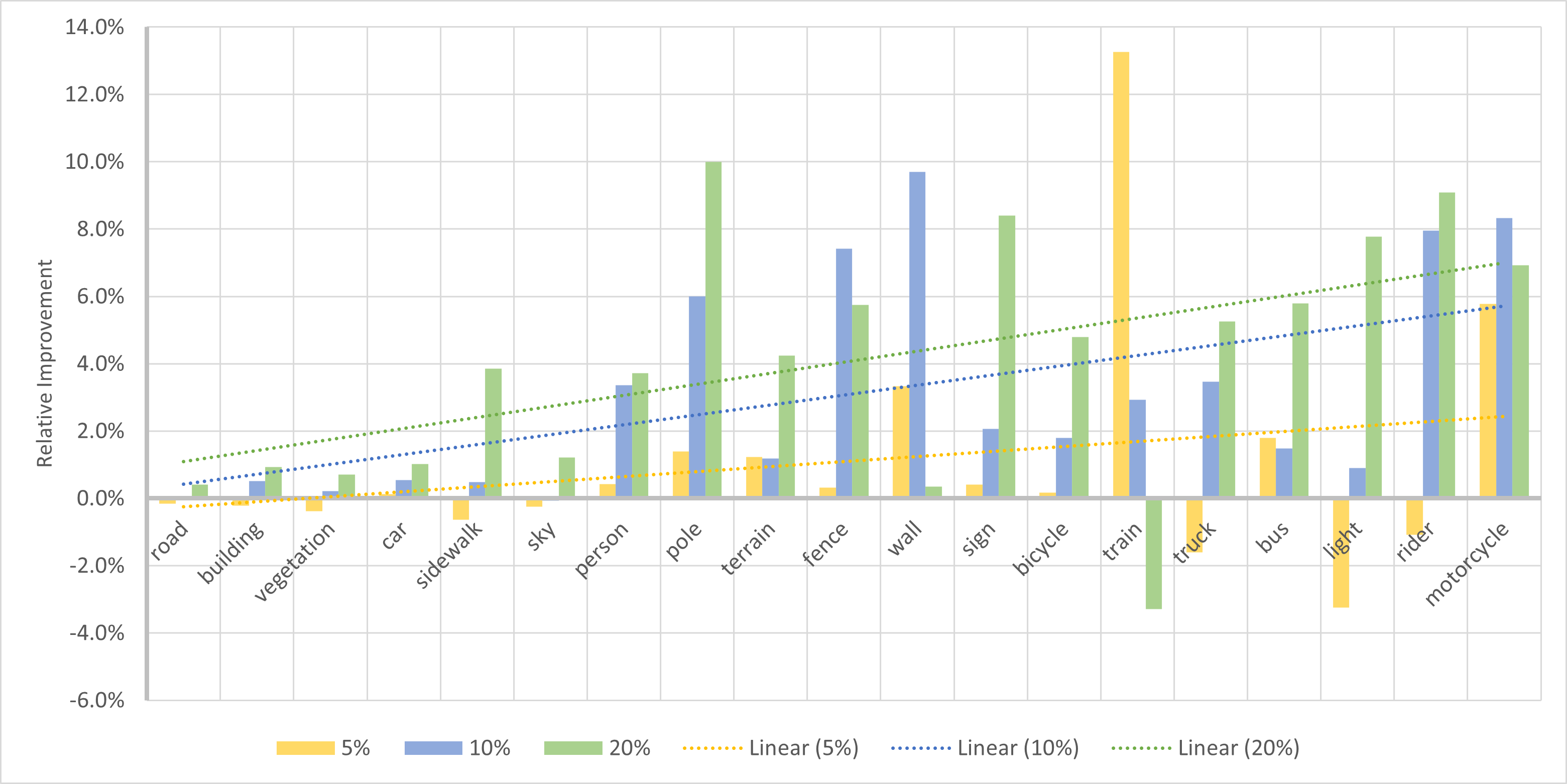}
    \caption{Relative improvements on each class (CBDA over RA for GTAV $\rightarrow$ Cityscapes) showing increased improvement for minority classes}
    \label{fig:minority_improv}
\end{figure}

To further show the performance differences of the minority classes, the graph in \cref{fig:minority_improv} shows the relative improvements of each class compared to the RA baseline with the same budget.
As in the results tables, the classes are in descending order by frequency.
The graph clearly shows the overall stable behavior for the top 4 classes, making up $77.7\%$ of the total pixels and the general increases for the minority classes.
There are a few outliers, such as \textit{train} and \textit{light}, with specific drops for one of the budgets.
The broad trends are represented by the linear trendlines for each of the budgets and they show a progressively increasing relative performance with decreasing class frequency.
These trends should be seen as qualitative and may not exactly follow a linear function, especially since the horizontal axis is categorical.
However, to verify that the trends are actually there, an average trendline for ten different random orders of the classes was calculated.
These trendlines had negligible slopes, indicating that the trendlines shown here are relative to a steady baseline.
The increasing performance of each budget for CBDA is still regarding the same budget for RA, so the large increase for CBDA at 20\% is with reference to the rather poorly performing baseline of RA at 20\%.
While 5\% has a smaller improvement, it is nonetheless performing better in absolute terms.

\newpage
\section{      } %
\label{sec:ablation_da_cb}

\subsection*{Dynamic Acquisition \& Class Balancing}

This ablation study studies the performance and application of \textit{Dynamic Acquisition} and \textit{Class Balancing} compared to the original RA baseline method.
The objective is to examine the differences between RA and DA, particularly when combined with class balancing.
As previously shown in \cref{fig:da_usage}, the extra flexibility of DA is being used; however, it is not clear if that provides a performance boost.

\begin{figure}[ht]
    \centering
    \begin{subfigure}[b]{0.48\textwidth}
        \centering
        \includegraphics[width=\textwidth]{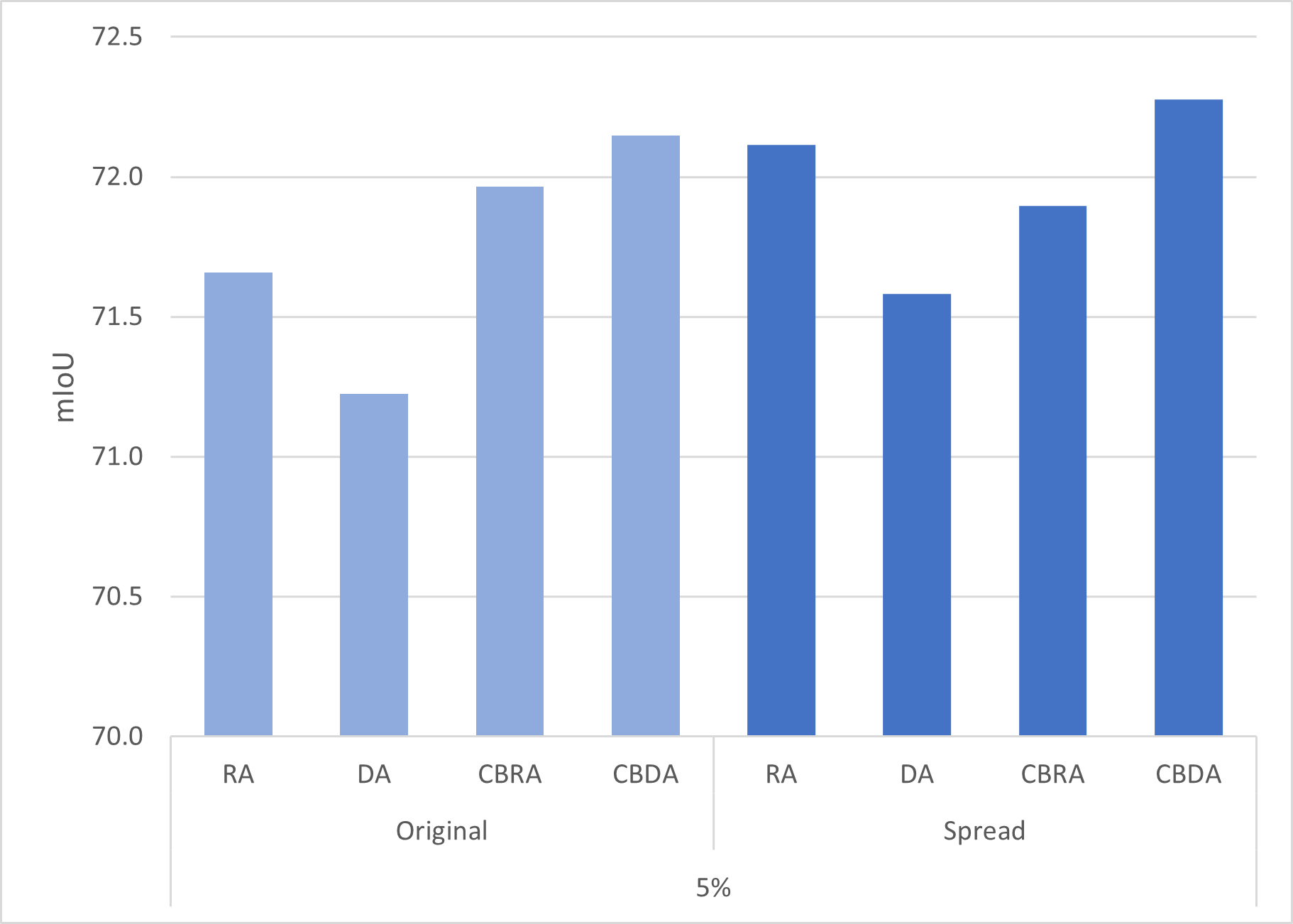}
    \end{subfigure}
    \hfill
    \begin{subfigure}[b]{0.48\textwidth}
        \centering
        \includegraphics[width=\textwidth]{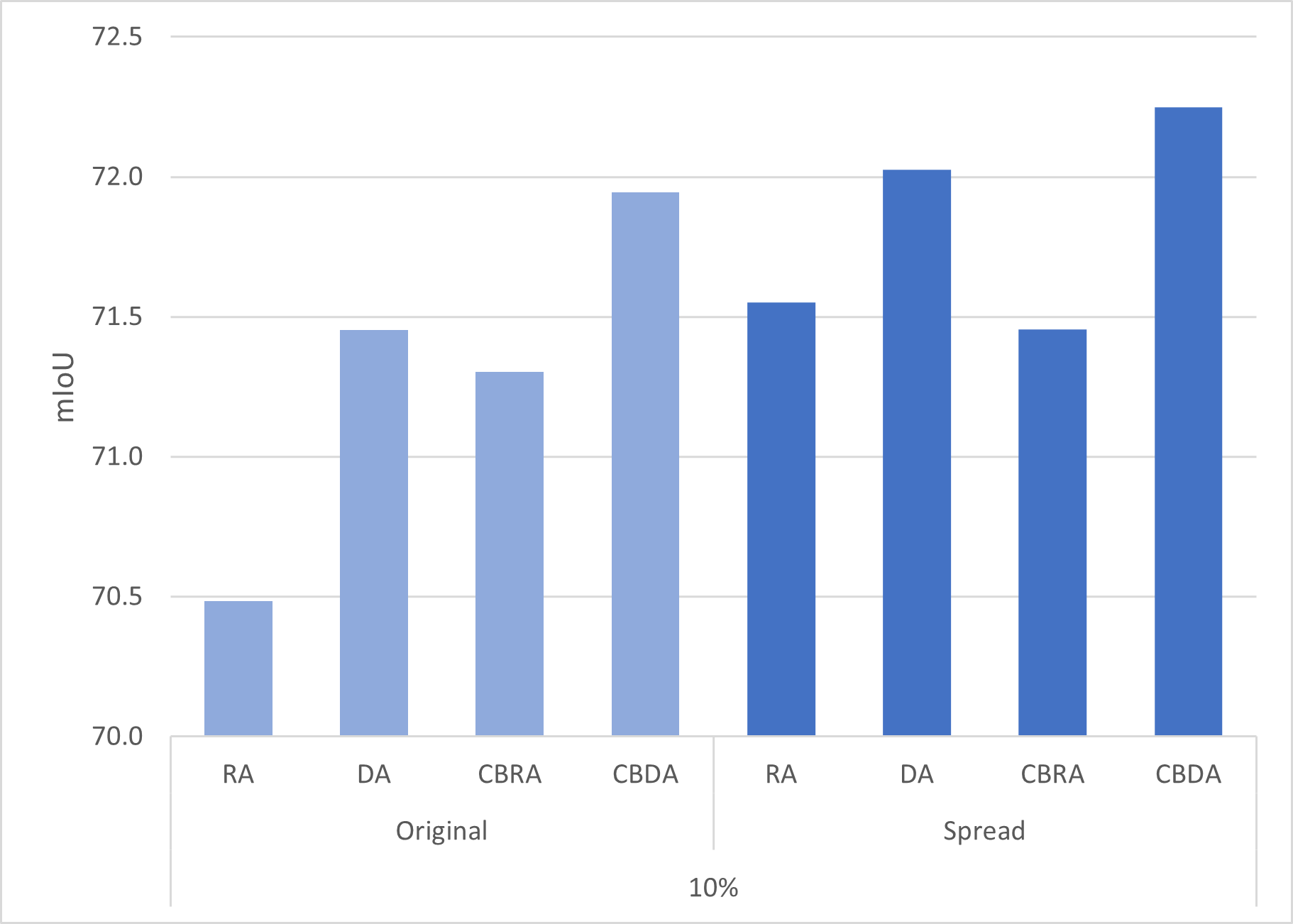}
    \end{subfigure}
    \caption{Ablation study comparing RA, DA, CBRA and CBDA}
    \label{fig:abl_cbda}
\end{figure}

\cref{fig:abl_cbda} shows the results of the ablation study.
While the results are generally reasonably consistent, runs using just DA tend to vary more than any other combinations, and the values shown here are of a \textit{typical} run.

In absolute terms, CBDA is the best-performing method compared to all others and for both budgets.
In the baseline paper active learning was performed at training iterations $[10000, 12000, 14000, 16000, 18000]$ - marked as \textit{Original}.
However, using iterations $[10000, 18000, 24000, 28000, 30000]$ were found to lead to better results and are marked here as \textit{Spread}. 
Important to note here is that \textit{spread} generally outperforms the \textit{original} setup and provides a reasonable boost in performance all by itself.

The most significant difference between the 5\% and 10\% AL budgets in terms of the relative performance of each method to the others is the improvement of class balancing vs the unweighted equivalent.
Except for CBRA performing slightly lower than RA  with \textit{spread} AL iterations, class balancing is an improvement.
An explanation for the slightly reduced score in the \textit{spread} setup can be that the class balancing setup can discourage certain high-scoring indices from being selected.
Using the \textit{original} iterations for active learning limited the amount of convergence between each AL iteration.
This, in turn, caused fewer representative pixels to be selected and led to a lower score; class balancing improved that score by reducing the imbalance, increasing the minority class IoU scores more than it cost in selecting indices with lower acquisition scores.
Perhaps with the \textit{spread} AL iteration setup, this trade-off is not worth it anymore, at least with the already limiting equal budget per image setup of RA.
The model has more iterations to converge more, meaning it can better select the pixels that should get an active label.
Limiting those pixels now may lead to a lower overall score, albeit likely still improving in some minority classes because of the more balanced selection method.

The equal budget per image is the most significant issue when pairing class balancing with the RA setup.
As previously outlined, an equal budget per image means that each class should have an equal number of pixels for each image.
Suppose a minority class is not represented in a large amount of those images. 
In that case, this method has challenges making it up from another image since the budget is also limited for that image.
The domain acquisition (DA) method was designed to eliminate that limitation and allow the model more flexibility in selecting which pixels to get labeled.
By itself, DA may be more flexible but does not guarantee a score improvement - on the contrary, for the 5\% budget, it significantly underperforms RA.
However, using the flexibility of DA by limiting some of its index selection with the class balancing weights uses the respective advantages to combat the respective disadvantages.
The flexibility of DA alone may be challenging for a not adequately converged model; class balancing limits the selection.
On the other hand, class balancing artificially reduces the acquisition scores and may cause some high-scoring indices not to be selected.
The extra flexibility of DA may ensure that enough high-scoring indices with appropriate classes are available for selection since it selects on the dataset scale and not the image scale.

\subsection*{Quantifying Imbalance}

Comparing the active label imbalance for the baseline showed a progressively increasing imbalance for larger budgets.
To quantify the imbalance of the active label, the KL divergence between the class distribution $Q_{class}$ and the uniform distribution $Q_{unif}$, scaled to between 0 and 1 using the maximal possible divergence, was calculated.
The formula for this \textit{imbalance score} is shown in \cref{eqn:imbalance_score}. 
The maximum possible divergence is between a one-hot distribution $Q_{OH}$ (all labels for a single class) and the uniform distribution.

\begin{equation}
    \label{eqn:imbalance_score}
    imbalance\_score = \frac{\KLnew{Q_{class}}{Q_{unif}}}{\KLnew{Q_{OH}}{Q_{unif}}} 
\end{equation}

\begin{figure}[ht]
    \centering
    \includegraphics[width=0.9\textwidth]{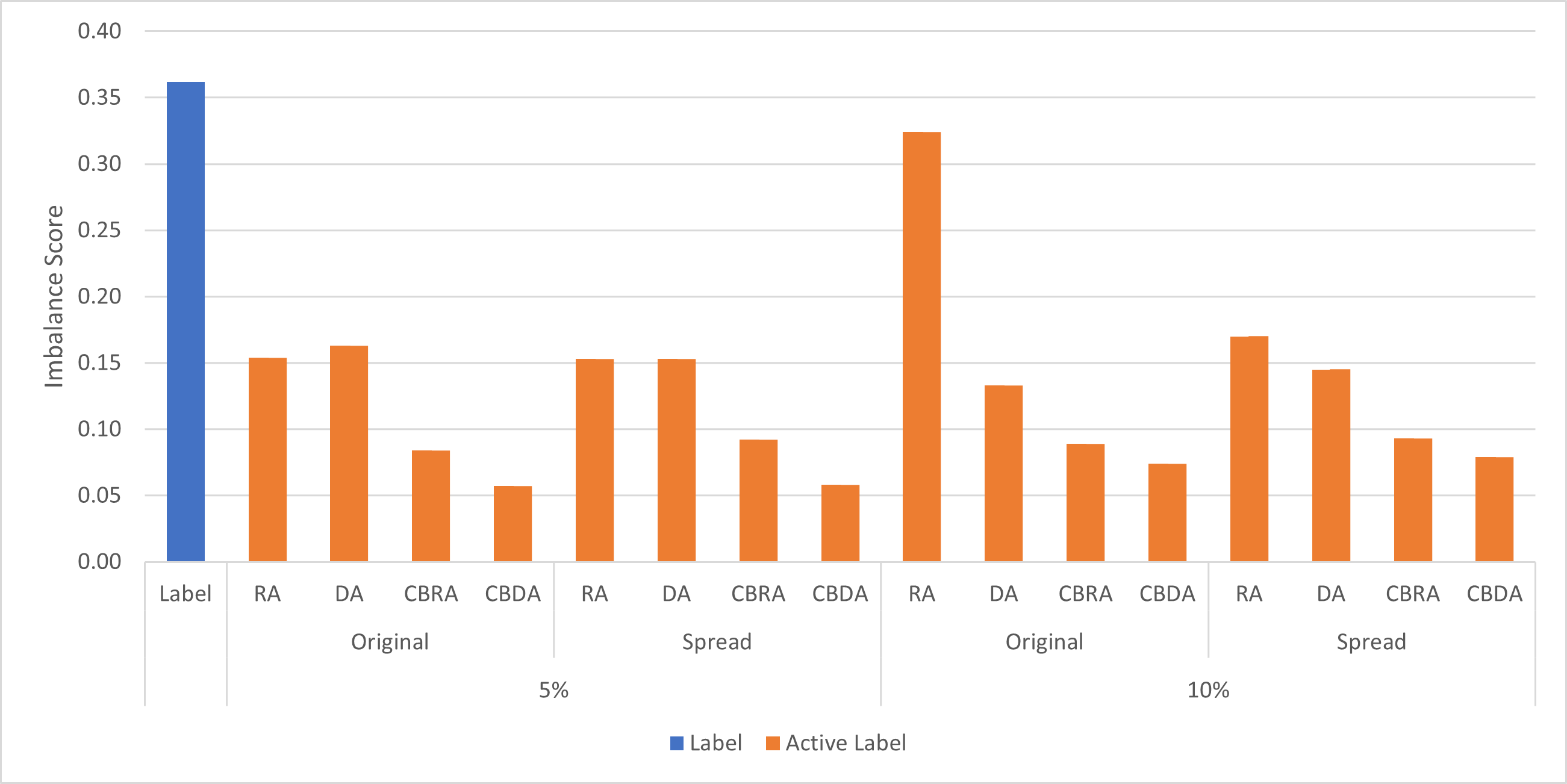}
    \caption{Imbalance scores for different setups showing lowest imbalance for methods incorporating CB}
    \label{fig:abl_cbda_imb_scores}
\end{figure}

\cref{fig:abl_cbda_imb_scores} shows the imbalance scores for the previous experiments.
The values should be taken in the context of the original RA runs.
In addition, the original imbalance score of an active label with a 20\% budget was 0.56, but no ablations for 20\% were performed.

First of all, this graph clearly shows the balancing effect of the class balancing methodology.
The selection imbalance is reduced from around $0.16$ and $0.33$ to less than $0.06$ and $0.08$ for 5\% and 10\%, respectively.
While class imbalance is undoubtedly not the only factor that affects the final score, this shows that the reduction in the class imbalance is correlated with a score improvement.
Even the score reduction from RA to DA for the 5\% budget matches the slight increase in imbalance caused by the switch.
The slight score decrease between RA and CBRA for the \textit{spread} AL iteration setup, however, is not.
As previously explained, the current assumption is that the class balancing causes less important pixels to be selected in the RA regime with a model that is better converged than the original.
This issue seems to be resolved when pairing class balancing with DA.

\end{document}